\pdfoutput=1

\documentclass[11pt]{article}

\usepackage[]{acl}

\usepackage{times}
\usepackage{latexsym}

\usepackage{colortbl}
\usepackage{amsthm}
\usepackage{amsfonts} 
\usepackage{url}   
\usepackage{float}
\usepackage{stfloats}
\usepackage{subfig}
\usepackage{enumitem}
\usepackage{ifthen}
\usepackage{color, xcolor}
\hypersetup{colorlinks=true, linkcolor=blue, citecolor=blue, anchorcolor=blue, urlcolor=blue}
\usepackage{thm-restate}
\usepackage{algorithm}
\usepackage{algorithmic}
\usepackage{enumitem}
\usepackage{multirow}
\usepackage{array}
\usepackage{microtype}
\usepackage{siunitx}
\usepackage{enumitem}
\usepackage{lineno}
\usepackage{tabularx} 
\usepackage{booktabs} 
\usepackage{setspace}
\usepackage{graphicx}
\usepackage{listings}
\usepackage{xcolor}
\usepackage{makecell}

\newlist{enuminline}{enumerate*}{1}
\setlist[enuminline,1]{label=\itshape\alph*\upshape)}

\newcommand{\lgh}[1]{\textcolor{red}{#1}}   

\usepackage[T1]{fontenc}

\usepackage[utf8]{inputenc}

\usepackage{microtype}

\makeatletter
\newcommand{\finalcopy}{\def\ifacl@finalcopy{\iftrue}}
\makeatother
\finalcopy
\title{Re-Search for The Truth: Multi-round Retrieval-augmented Large Language Models are Strong Fake News Detectors}

\author{
    \begin{tabular}{c}
    Guanghua Li$^1$ \quad Wensheng Lu$^1$ \quad Wei Zhang$^1$ \quad Defu Lian$^2$ \\
    Kezhong Lu$^1$ \quad Rui Mao$^1$ \quad Kai Shu$^3$ \quad Hao Liao$^1$
    \end{tabular}
    \\
    \small
    \begin{tabular}{c}
    College of Computer Science and Software Engineering, Shenzhen University, China$^1$ \\
    School of Computer Science and Technology, University of Science and Technology of China$^2$ \\
    Illinois Institute of Technology$^3$ \\
    \end{tabular}
    \\
    \small
    \begin{tabular}{c}
    \texttt{\{2210275050, 2210273060, 2210275010\}@email.szu.edu.cn} \quad \\
     \texttt{liandefu@ustc.edu.cn} \quad \texttt{\{kzlu, mao\}@szu.edu.cn}\\
    \texttt{kshu@iit.edu} \quad \texttt{haoliao@szu.edu.cn}
    \end{tabular}
}

\begin{document}
\maketitle
\begin{abstract}
The proliferation of fake news has had far-reaching implications on politics, the economy, and society at large. While Fake news detection methods have been employed to mitigate this issue, they primarily depend on two essential elements: the quality and relevance of the evidence, and the effectiveness of the verdict prediction mechanism. Traditional methods, which often source information from static repositories like Wikipedia, are limited by outdated or incomplete data, particularly for emerging or rare claims. Large Language Models (LLMs), known for their remarkable reasoning and generative capabilities, introduce a new frontier for fake news detection. However, like traditional methods, LLM-based solutions also grapple with the limitations of stale and long-tail knowledge. Additionally, retrieval-enhanced LLMs frequently struggle with issues such as low-quality evidence retrieval and context length constraints. To address these challenges, we introduce a novel, retrieval-augmented LLMs framework—the first of its kind to automatically and strategically extract key evidence from web sources for claim verification. Employing a multi-round retrieval strategy, our framework ensures the acquisition of sufficient, relevant evidence, thereby enhancing performance. Comprehensive experiments across three real-world datasets validate the framework's superiority over existing methods. Importantly, our model not only delivers accurate verdicts but also offers human-readable explanations to improve result interpretability. 
\end{abstract}

\section{Introduction}

The escalation of fake news poses a severe threat, dwarfing extensive efforts to mitigate its impact on political, economic, and social landscapes~\cite{West2020}. Fake news detection approaches to combat this issue generally fall into three categories: content-based~\cite{DBLP:journals/csur/ZhouZ20, DBLP:journals/ijon/CapuanoFLN23}, evidence-based~\cite{DBLP:conf/coling/KotonyaT20, DBLP:conf/www/MinRBXZHA22}, and social context-based methods~\cite{DBLP:journals/jiat/CollinsHNH21, DBLP:conf/nips/GroverAA022}.


However, existing methods~\cite{zhou2020survey, DBLP:journals/ipm/ZhangG20} are typically tailored to specific datasets, thereby inherently constraining their scalability, transferability, and robustness.
In light of these constraints, there arises an imperative for the development of a more versatile model that can efficiently detect fake news in a zero-shot or few-shot learning manner.


\begin{figure}[!t]
    \centering
    \includegraphics[width=0.48\textwidth]{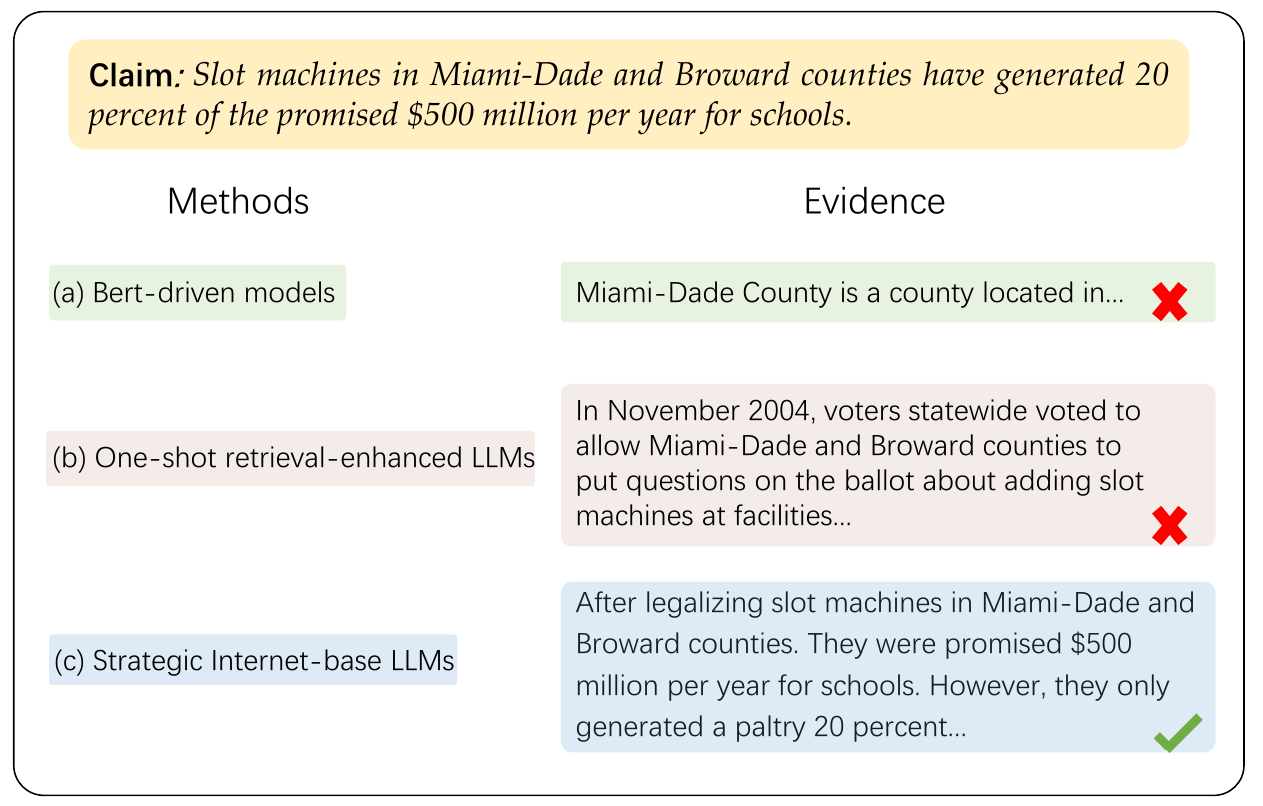}
    \caption{A motivating example of our model. (a) Bert-driven methods: up-to-date evidence cannot be retrieved. (b) One-shot retrieval-enhanced LLMs: only partial evidence can be retrieved. (c) Strategic Internet-based LLMs: multi-round retrieval of evidence from the Internet facilitates more comprehensive and accurate assessments.}
    \label{fig:intro}
\end{figure}

Large Language Models (LLMs) have shown remarkable capabilities across various applications~\cite{DBLP:journals/tmlr/WeiTBRZBYBZMCHVLDF22}. Current methodologies utilizing Retrieval-Augmented Generation (RAG) and Large Language Models (LLMs) often depend on specific databases such as Wikipedia or employ a simple one-step retrieval process~\cite{DBLP:journals/jmlr/Gautier23, 10.5555/3524938.3525306}. However, in the context of real-world fake news detection, there are significant systemic challenges that necessitate more sophisticated solutions. These challenges encompass the growing issue of AI-generated disinformation, the limitations inherent in depending on a limited number of data sources, the obstacles of ensuring real-time updates in a constantly changing news environment, and the long-tail effect where rare or niche false information may remain undetected~\cite{DBLP:journals/corr/abs-2311-05656}. In response to these obstacles, we propose an innovative multi-round LLM-based RAG framework.

We introduce \textbf{STEEL} (\underline{\textbf{ST}}rategic r\underline{\textbf{E}}trieval \underline{\textbf{E}}nhanced with \underline{\textbf{L}}arge Language Model), a comprehensive, automated framework for fake news detection that combines ease-of-use and interpretability. Our framework leverages the reasoning and uncertainty estimation capabilities of LLMs, offering more robust evidence retrieval. It also sidesteps the limitations of relying on a solitary predefined corpus by sourcing evidence directly from the expansive Internet. As illustrated in Figure \ref{fig:intro}, STEEL employs an adaptive multi-round retrieval process, using a Large Language Model to generate targeted queries for missing information when initial evidence is insufficient. In addition, it can sharpen the focus of subsequent retrievals and save crucial evidence already obtained for the next judgment.

In this work, we make the following contributions.
\begin{itemize}
    \item We propose a novel framework, STEEL, for automatic fake news detection through strategic Internet-based evidence retrieval. To the best of our knowledge, this is the first framework that leverages LLMs for fake news detection via strategic evidence retrieval from the Internet. 
    \item We provide an open-source implementation that is designed for out-of-the-box use, eliminating the need for complicated data processing or model training. 
    \item Extensive experiments on three real-world datasets demonstrate that STEEL outperforms state-of-the-art methods in both prediction and interpretability.   
\end{itemize}
\section{Related Work}

\subsection{RAG LLMs}

The retrieval-augmented language model assists text generation by retrieving relevant documents from a vast external knowledge base~\cite{DBLP:journals/corr/abs-2112-09332}. 
This combats long-tail, outdated knowledge, and hallucination issues~\cite{DBLP:conf/icml/KandpalDRWR23}. Recent work has shown that retrieving additional information can improve performance on a variety of downstream tasks~\cite{DBLP:conf/iclr/YaoZYDSN023}, including open-domain Q\&A, fact-checking, fact completion, long-form Q\&A, Wikipedia article generation, and fake news detection~\cite{DBLP:conf/iclr/0002IWXJ000023, DBLP:conf/icml/GuuLTPC20, DBLP:conf/acm/AsaiMZC23, DBLP:conf/cikm/WuLDXH23, DBLP:conf/emnlp/WangS23a}.

STEEL differs notably from other retrieval methods in the RAG+LLM framework, like FLARE~\cite{DBLP:conf/emnlp/JiangXGSLDYCN23}, Replug~\cite{DBLP:journals/corr/abs-2301-12652}, ProgramFC~\cite{DBLP:conf/acl/PanWLLWKN23}, and SKR~\cite{DBLP:conf/emnlp/WangLSL23}. While FLARE, ProgramFC, and SKR focus mainly on text blocks, Replug on documents, STEEL retrieves both documents and text blocks. Unlike methods relying on Wikis, STEEL uses the Internet as its source. It shares context-based retrieval timing with other methods but introduces active search features, including LLM feedback utilization and answer verification, enhancing its flexibility and depth in retrieval tasks within the RAG+LLM framework.


\subsection{Natural Language Inference LLMs}

Natural Language Inference (NLI) is used to predict the logical connection between the claim and the provided evidence.
Recent studies have made strides in enhancing LLMs' reasoning. Chain of Thought \cite{DBLP:conf/nips/Wei0SBIXCLZ22} achieved significant improvements with simple prompt modifications. ReAct \cite{DBLP:conf/iclr/YaoZYDSN023} integrates reasoning and acting capabilities in LLMs for better performance in tasks requiring complex reasoning. Tree of Thoughts \cite{DBLP:journals/corr/abs-2305-10601} enables deliberate decision-making in LLMs by exploring reasoning paths and facilitating self-evaluation. In contrast, our work focuses on evidence-retrieval strategies for news verification.
Currently, main application paradigms can be divided into: Prompting~\cite{ram-etal-2023-context}, Fine-tuning~\cite{DBLP:conf/icml/BorgeaudMHCRM0L22}, and Reinforcement learning~\cite{DBLP:conf/kdd/LiuLYXZDZDT23}. Existing industrial solutions like NEW BINGBING~\footnote{\url{https://www.bing.com/new}} and Perplexity.ai~\footnote{\url{https://www.perplexity.ai/}} integrate LLMs with search engines for performance but aren't optimized for fake news detection. In this task, evidence quality is crucial due to LLM input length limits. STEEL addresses this by using LLM feedback and multi-round evidence retrieval.
\section{Methods}
In this section, we present our model, STEEL. The input of this method consists of a claim $C$. Initially, a set of relevant evidence $E_v = \{E_1,E_2,E_3,...\}$ are retrieved from the Internet. Subsequently, LLMs evaluate the sufficiency of the gathered evidence. If the evidence is deemed adequate, the results will be output promptly. Otherwise, the search for additional evidence continues. To construct an affordable, ready-to-use framework, we leverage the APIs (Application Interfaces) of leading AI (Artificial Intelligence) companies. Specifically, we utilize BING Search for web evidence retrieval and OPENAI's GPT-3.5-turbo ~\cite{openai2022chatgpt} for verification. The output is the prediction of this claim $\hat{y}$, along with explanatory text $E_x=L(C, E_v)$. Here, $L$ refers to the LLMs responsible for generating the output. $y$ is a binary classification, where $y\in\{true, false\}$ indicates the assessment of the news claims as true or false. 

\begin{figure*}[t]
    \centering
    \includegraphics[width=\textwidth]{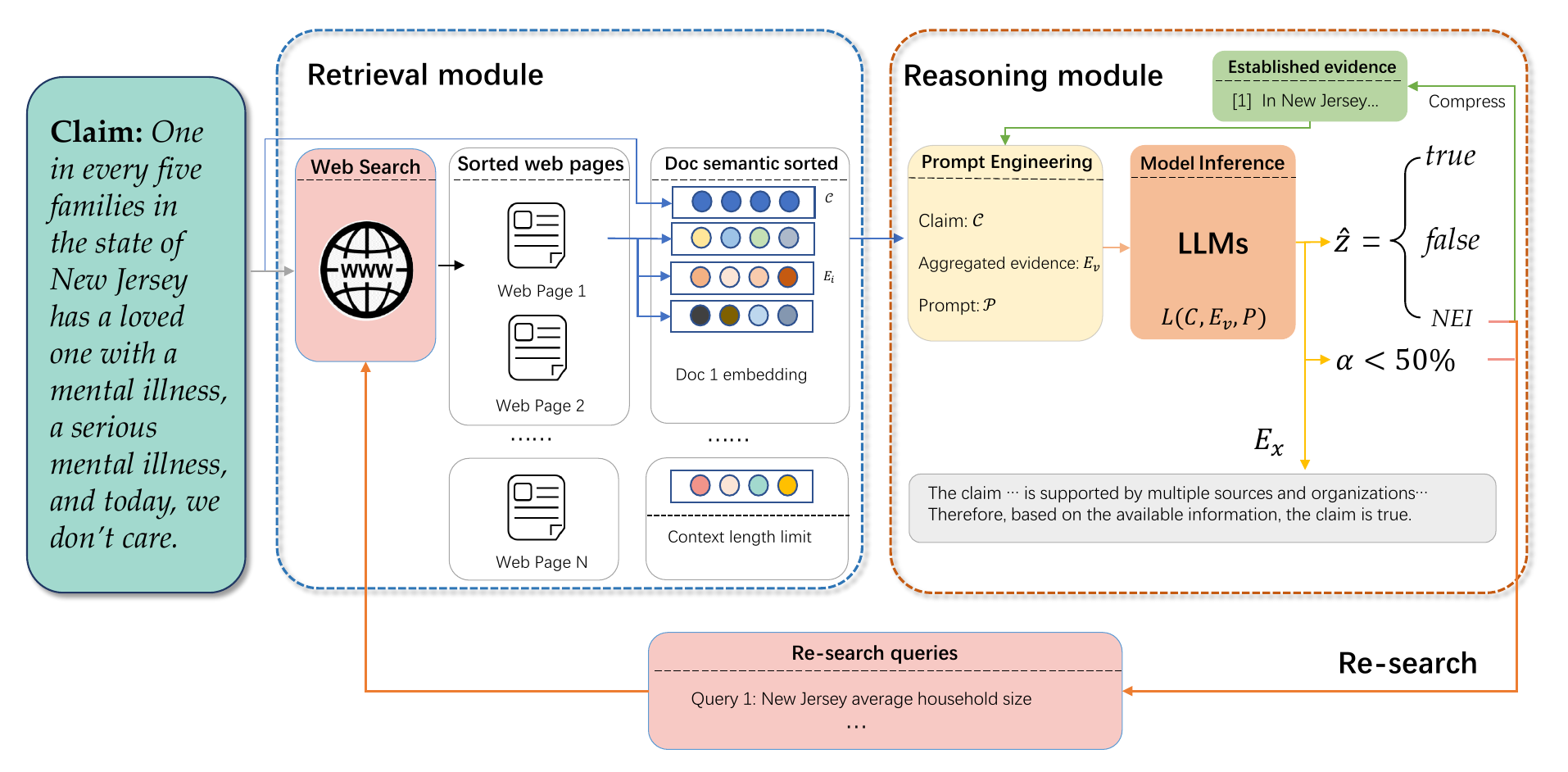}
    \caption{The overview of the STEEL framework. Our framework unfolds in three parts: (a) Retrieval module. Use claim or updated queries to search for evidence via the search engine, sort and select based on the similarity between the searched documents and paragraphs of the claim. (b) Reasoning module. Feed the obtained evidence and established evidence to LLMs via carefully designed prompts, and LLMs will reason and output one of the three situations "true, false, or NEI (Not Enough Information)" with confidence levels. Even when the output is "NEI", LLMs will compress the newly obtained information to the pool of established evidence for subsequent search. (c) Re-search mechanism. Re-search for more evidence when the output is "NEI" or the confidence level is below $50\%$. We use LLMs to generate "updated queries" to improve the quality of retrieval evidence.
    }
    \label{fig:model_overview}
\end{figure*}

As shown in Figure \ref{fig:model_overview}, our model mainly comprises two key components: a retrieval module and a reasoning module. These two modules are integrated within the overarching framework of the re-search mechanism. 

\subsection{Retrieval Module}

Unlike prior studies that separate web retrieval and semantic retrieval, we integrate both stages. A claim $C$ is first processed by a web retrieval API to obtain document links $U_s$ containing pertinent evidence. Typically, 10 links are retrieved; however, due to constraints imposed by the context length of Large Language Models (LLMs), not all links are utilized. 


For source credibility, we implement a basic filtering mechanism. Based on previous research~\cite{DBLP:conf/www/Papadogiannakis23}, we use a list of $1,044$ known fake news websites as a filter, discarding any matches during web search.

The documents retrieved online are initially organized based on the relevance algorithm of the search engine, with the document deemed most relevant positioned at the top of the list. Our analytical process adheres to the sequence of this sorted list, beginning with the first document. Specifically, our approach involves assessing whether the length of the top-ranked document exceeds our predefined limit determined by the LLM's context length. If it does, we employ semantic retrieval techniques to extract highly similar fragments from the document. Conversely, if the length is within acceptable limits, we utilize the entire document and then sequentially examine the second-ranked document, continuing this process until we reach the maximum allowable context length. By this, we strive to gather a comprehensive array of relevant evidence while maintaining the integrity of the information retrieved.

\subsection{Reasoning Module}

The relevant pieces of evidence $E_v$ retrieved from the web are then aggregated into prompts and fed into the LLMs for inference. LLMs can make decisions based on given
evidence, including deciding if they need to re-search, case can be seem at Figure \ref{fig:hypo}.
Essentially, the prompt instructs LLM to assess the claim based on the retrieved evidence and output responses, which are classified into three categories - true, false, and NEI (Not Enough Information). Explanations of the responses are provided based on the sufficiency of the retrieved evidence. For "NEI", "Established evidence" and "Updated queries" will be output for further evidence collection. "Established evidence" is the compression of this evidence for the next judgment. "Updated queries" are the queries for subsequent web page retrieval, with the purpose of incrementally obtaining evidence. Prompts utilized here can be seen in listing \ref{lst:step_check}. 
To mitigate consistency issues, we incorporate a confidence level for each answer, along with aggregated new and established evidence for subsequent assessment.
The third is aggregated evidence of $E_v$ obtained after retrieval and "Established evidence" in the previous cycle.

\begin{equation}
    \hat{y},E_x=L(C,E_v,P)
\end{equation}


To address inconsistent answers~\cite{DBLP:conf/nips/YeD22} and hallucinations problem, some previous work~\cite{DBLP:journals/corr/abs-2306-13063, DBLP:conf/iclr/0002WSLCNCZ23} exploits the self-consistence and self-judgment approaches, enabling the LLMs to produce confidence scores within the range of $[0, 100\%]$. Nonetheless, it has been observed that contemporary LLMs often exhibit a tendency toward overconfidence~\cite{DBLP:journals/corr/abs-2307-05300, DBLP:journals/corr/abs-2306-13063}. To counteract this phenomenon, we introduce an over-confidence coefficient within the range of $[0, 1]$. The final confidence score is adjusted by multiplying it with this coefficient. 
When the final Confidence falls below $50\%$, the model is instructed to proceed to the next iteration.
\begin{equation}\label{equ:discount}
    \alpha=\beta* Conf
\end{equation}
In equation \ref{equ:discount}, $\alpha$ denotes the final confidence score, $Conf$ represents the initial confidence score provided by the LLMs, and $\beta$ represents the over-confidence coefficient.

\subsection{Re-Search Mechanism} 
\begin{figure}[!pt]
    \centering
    \includegraphics[width=0.48\textwidth]{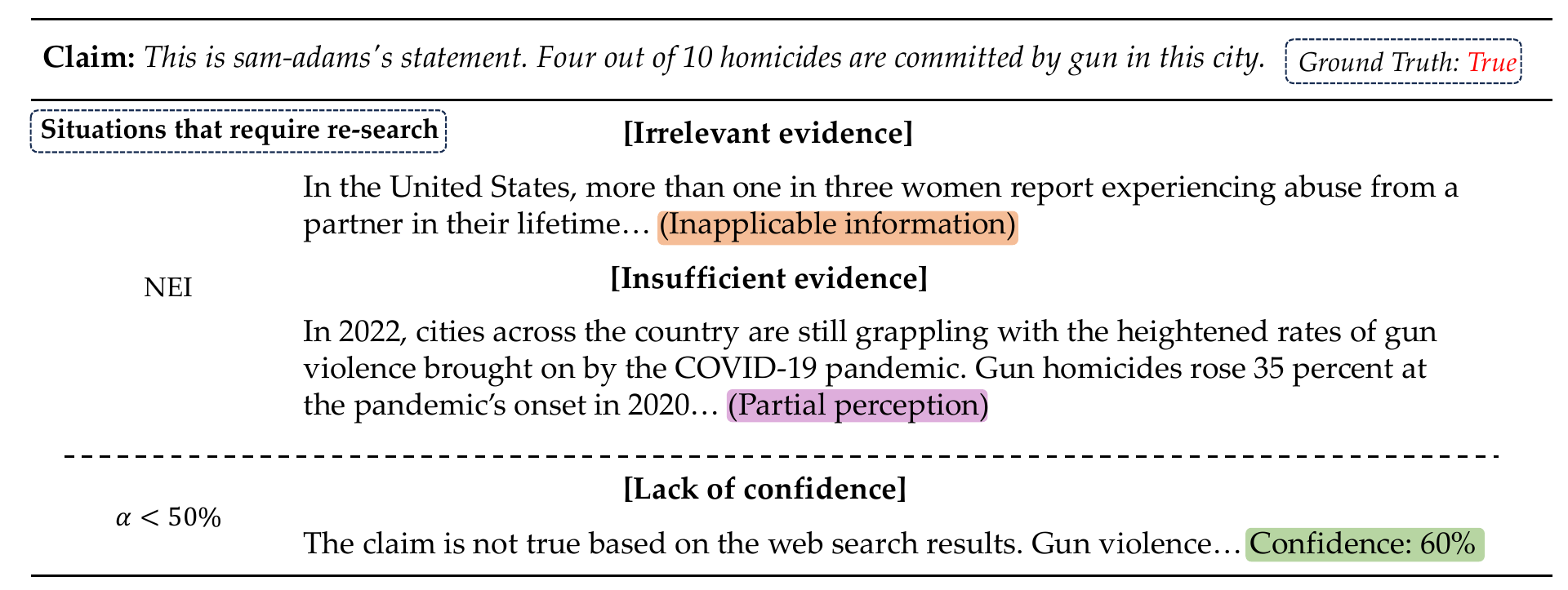}
    \caption{Situations necessitating re-search: \emph{Irrelevant Evidence} denotes evidence unrelated to the query or claim. \emph{Insufficient Evidence} indicates inadequate evidence for reaching a valid conclusion. \emph{Lack of Confidence} signifies uncertainty or low confidence in the conclusion's accuracy based on evidence.}
    \label{fig:method_rs}
\end{figure}

As illustrated in Figure \ref{fig:method_rs}, the re-search is triggered under certain conditions. This feature ensures a more robust and exhaustive gathering of evidence, enhancing the method's reliability and performance.

Upon meeting a re-search condition, the model kicks off a systematic process. First, it consolidates the evidence gathered from the initial search, appending it to an "established evidence" pool for future reference. Next, the model formulates a set of "updated queries" aimed at obtaining additional relevant evidence. This iterative approach ensures a gradual accumulation of evidence, thereby enhancing the model’s ability to discern the veracity of news.

Regarding the rationale behind our choice of re-search over alternative methods that appear to enhance retrieval quality, such as query-dependent techniques or search engineering, a detailed explanation will be provided in section~\ref{Internet Search Comparison Study}. 

Consequently, when LLMs determine that the current evidence set is inadequate for a reliable judgment on the claim at hand, it signals this by outputting "NEI". This output serves as a trigger for the model to advance to a subsequent iterative search. The mechanics behind this intermediate step are further detailed in Equation \ref{equ:estimation}.

\begin{equation}\label{equ:estimation}
    \hat{z},E_x,\alpha=L(C,E_v,P)
\end{equation}
where $\hat{z}\in\{true,false,NEI\}$ and NEI stands for Not Enough Information.
\section{Experiments}\label{sec: evaluation}
In this section, we conduct experiments to evaluate the efficacy of our STEEL model in multiple angles. We focus on three main aspects: the efficiency of evidence retrieval in identifying fake news, the role of the re-search mechanism in bolstering detection accuracy, and the influence of varying retrieval steps and prompts on the model's performance.

\subsection{Experiments Setup}

\paragraph{\textbf{Datasets}}
To evaluate the performance of STEEL, we conduct extensive experiments on three real-world datasets, comprising two English datasets (LIAR\footnote{\href{https://www.cs.ucsb.edu/~william/data/liar_dataset.zip}{LIAR:https://www.cs.ucsb.edu/~william/data/liar\_dataset.zip}} and PolitiFact\footnote{\href{https://www.politifact.com/}{PolitiFact: https://www.politifact.com/}}) and one Chinese dataset (CHEF\footnote{\href{https://github.com/THU-BPM/CHEF}{CHEF: https://github.com/THU-BPM/CHEF}}). 
The news in LIAR and PolitiFact are categorized into two distinct classes: real and fake news. The datasets were preprocessed to maintain their original meaning while fitting the task at hand, with key statistics outlined in Table \ref{tab:datasets}. 

\begin{table}
  \centering
  \small

  \begin{tabular}{lccc}
    \toprule
     & LIAR & CHEF & POLITIFACT\\
    \midrule
    \#Real News & $9{,}252$ & $3{,}543$ & $399$ \\
    \#Fake News & $3{,}555$ & $5{,}015$ & $345$ \\
    \#Total & $12{,}807$ & $8{,}558$ & $744$\\
  \bottomrule
    \end{tabular}
  \caption{Statistics of the datasets.}
  \label{tab:datasets}
\end{table}



\paragraph{\textbf{Baselines}}
We compare our STEEL with $11$ baselines, which can be divided into two groups: 

The first group (G1) is classical and recent advanced evidence-based methods. G1 contains seven baselines: \textbf{DeClarE} (EMNLP’18) \cite{2018-declare}, \textbf{HAN} (ACL'19) \cite{DBLP:conf/acl/MaGJW19}, \textbf{EHIAN} (IJCAI’20)~\cite{2020-ehian}, \textbf{MAC} (ACL’21)~\cite{2021-mac}, \textbf{GET} (WWW’22)~\cite{2022-get}, \textbf{MUSER} (KDD'23)~\cite{DBLP:conf/kdd/LiaoPHZLSX23} and \textbf{ReRead} 
(SIGIR'23)~\cite{DBLP:conf/sigir/HuHGWY23}.

The second group (G2) encompasses methods based on LLMs, either with or without a retrieval component. This group includes four methods: \textbf{GPT-3.5-turbo}~\cite{openai2022chatgpt}, \textbf{Vicuna-7B}~\cite{chiang2023vicuna}, \textbf{WEBGLM} (KDD'23)~\cite{DBLP:conf/kdd/LiuLYXZDZDT23}and \textbf{ProgramFC} (ACL'23)~\cite{DBLP:conf/acl/PanWLLWKN23}.
For a detailed description of the baseline models, please refer to the Appendix \ref{app: baselines}.

\begin{table}[t]


\centering
\tabcolsep=3pt
\resizebox{0.48\textwidth}{!}{%
\begin{tabular}{cccccccccc}
\toprule[1pt]
\multicolumn{2}{c}{\multirow{2}{*}{Method}} & \multicolumn{8}{c}{LIAR}  \\ 
\cline{3-10}
 & & F1-Ma & F1-Mi & F1-T &  P-T & R-T & F1-F & P-F & R-F \\ 
\hline

\multirow{7}{*}{G1}   
&DeClarE & 0.573 & 0.571 & 0.531 & 0.550 & 0.546 & 0.619 & 0.587 & 0.597 \\
&HAN & 0.588 & 0.591 & 0.563 & 0.545 & 0.532 & 0.606 & 0.618 & 0.611 \\
&EHIAN & 0.591 & 0.593 & 0.559 & 0.543 & 0.548 & 0.630 & 0.603 & 0.617 \\
&MAC & 0.603 & 0.601 & 0.562 & 0.558 & 0.567 & 0.625 & 0.623 & 0.621 \\
&GET & 0.614 & 0.610 & 0.572 & 0.567 & 0.579 & 0.641 & 0.654 & 0.632 \\
&MUSER & 0.645 & 0.642 & 0.647 & 0.640 &  0.654 & 0.643 & 0.650 & 0.636 \\
&ReRead & 0.611 & 0.615 & 0.587 & 0.581 & 0.596 & 0.633 & 0.628 & 0.626 \\
\hline

\multirow{4}{*}{G2} 
& GPT-3.5-turbo & 0.563 & 0.541 & 0.559 & 0.572 & 0.567 & 0.555 & 0.564 & 0.560 \\
& Vicuna-7B & 0.528 & 0.535 & 0.521 & 0.543 & 0.552 & 0.519 & 0.538 & 0.526  \\
& WEBGLM-2B & 0.601 & 0.597 & 0.558 & 0.563 & 0.571 & 0.622 & 0.604 & 0.618  \\
& ProgramFC & 0.631 & 0.613 & 0.637 & 0.607 & 0.639 & 0.625 & 0.611 & 0.628 \\

\cline{2-10}
& \textbf{STEEL}  & \textbf{0.714*} & \textbf{0.689*} &\textbf{0.685*} & \textbf{0.680*} & \textbf{0.691*} &  \textbf{0.743*} & \textbf{0.725*} & \textbf{0.752*} \\
\bottomrule[1pt]
\end{tabular}%
}
\caption{Performance comparison on LIAR of our model w.r.t. baselines. The experiment was repeated 5 times with average results calculated. "F1-Ma" and "F1-Mi" denote F1-Macro and F1-Micro metrics. "-T" represents "True News as Positive," and "-F" denotes "Fake News as Positive" for precision and recall calculations. Statistically significant test ($P < 0.05$) performed on 5 dataset splits. Superior outcomes are highlighted in bold, and statistically significant improvements are indicated by *.}
\label{tab:performance_liar}
\end{table}

\begin{table}[!t]
\centering
\tabcolsep=3pt
\resizebox{0.48\textwidth}{!}{%
\begin{tabular}{cccccccccc}
\toprule[1pt]
\multicolumn{2}{c}{\multirow{2}{*}{Method}} & \multicolumn{8}{c}{CHEF}  \\ 
\cline{3-10}
 & & F1-Ma & F1-Mi & F1-T &  P-T & R-T & F1-F & P-F & R-F \\ 
\hline

\multirow{7}{*}{G1}   
&DeClarE & 0.589 & 0.581 & 0.637 & 0.583 & 0.625 & 0.568 & 0.544 & 0.581 \\
&HAN & 0.557 & 0.543 & 0.581 & 0.533 & 0.574 & 0.541 & 0.532 & 0.558 \\
&EHIAN & 0.600 & 0.571 & 0.621 & 0.583 & 0.628 & 0.577 & 0.516 & 0.586 \\
&MAC & 0.583 & 0.574 & 0.601 & 0.557 & 0.619 & 0.563 & 0.537 & 0.589 \\
&GET & 0.602 & 0.588 & 0.623 & 0.585 & 0.630 & 0.556 & 0.582 & 0.574 \\
&MUSER & 0.612 & 0.607 & 0.641 & 0.603 & 0.658 & 0.566 & 0.631 & 0.591 \\
&ReRead & 0.719 & 0.705 & 0.762 & 0.826 & 0.706 & 0.655 & 0.645 & 0.704 \\
\hline

\multirow{4}{*}{G2} 
& GPT-3.5-turbo & 0.574 & 0.586 & 0.567 & 0.571 & 0.595 & 0.583 & 0.579 & 0.591 \\
& Vicuna-7B & 0.519 & 0.513 & 0.509 & 0.538 & 0.531 & 0.522 & 0.518 & 0.525 \\
& WEBGLM-2B & 0.632 & 0.597 & 0.558 & 0.563 & 0.571 & 0.611 & 0.604 & 0.618 \\
& ProgramFC & 0.708 & 0.694 & 0.751 & 0.723 & 0.697 & 0.665 & 0.642 & 0.683 \\

\cline{2-10}
& \textbf{STEEL}  & \textbf{0.793*} & \textbf{0.781*} &\textbf{0.818*} & \textbf{0.850*} & \textbf{0.772*} &  \textbf{0.768*} & \textbf{0.725*} & \textbf{0.784*} \\
\bottomrule[1pt]
\end{tabular}
}
\caption{Performance comparison on CHEF of our model w.r.t. baselines.}
\label{tab:performance_chef}
\end{table}

\begin{table}[!t]

\centering
\tabcolsep=3pt
\resizebox{0.48\textwidth}{!}{%
\begin{tabular}{cccccccccc}
\toprule[1pt]
\multicolumn{2}{c}{\multirow{2}{*}{Method}} & \multicolumn{8}{c}{PolitiFact}  \\ 
\cline{3-10}
 & & F1-Ma & F1-Mi & F1-T &  P-T & R-T & F1-F & P-F & R-F \\ 
\hline

\multirow{7}{*}{G1}   
&DeClarE & 0.654 & 0.651 & 0.656 & 0.689 & 0.673 & 0.651 & 0.613 & 0.664 \\
&HAN & 0.661 & 0.660 & 0.679 & 0.676 & 0.682 & 0.643 & 0.650 & 0.637 \\
&EHIAN & 0.664 & 0.663 & 0.674 & 0.680 & 0.651 & 0.650 & 0.628 & 0.627 \\
&MAC & 0.678 & 0.675 & 0.700 & 0.695 & 0.704 & 0.653 & 0.655 & 0.645 \\
&GET & 0.694 & 0.692 & 0.725 & 0.712 & 0.770 & 0.669 & 0.720 & 0.665 \\
&MUSER & 0.732 & 0.729 & 0.757 & 0.735 & 0.780 & 0.702 & 0.728 & 0.681 \\
&ReRead & 0.681 & 0.693 & 0.714 & 0.711 & 0.755 & 0.688 & 0.718 & 0.699 \\
\hline

\multirow{4}{*}{G2} 
& GPT-3.5-turbo & 0.567 & 0.553 & 0.570 & 0.557 & 0.561 & 0.559 & 0.562 & 0.573  \\
& Vicuna-7B & 0.522 & 0.515 & 0.529 & 0.531 & 0.526 & 0.518 & 0.520 & 0.519  \\
& WEBGLM-2B & 0.628  &0.633  & 0.601 & 0.617 & 0.639 & 0.612 & 0.660 & 0.626  \\
& ProgramFC & 0.684 & 0.678 & 0.733 & 0.725 & 0.741 & 0.635 & 0.622 & 0.643\\

\cline{2-10}
& \textbf{STEEL}  & \textbf{0.751*} & \textbf{0.753*} &\textbf{0.780*} & \textbf{0.749*} & \textbf{0.787*} &  \textbf{0.722*} & \textbf{0.745*} & \textbf{0.724*} \\
\bottomrule[1pt]
\end{tabular}
}
\caption{Performance comparison on Politifact of our model w.r.t. baselines. }
\label{tab:performance_politifact}
\end{table}

\paragraph{Implementation details} Since our model does 
not require a training set, we utilize all the data as 
a test set. This approach is also applied to all the 
datasets we use. In our method, the hyperparameter $\beta$ is set to $0.7$. For the LLMs, we set the temperature at $0$, top-p at $0.75$, and limit prompt tokens to $4,096$. Hyperparameters for the baseline methods are aligned with those detailed in the respective papers and key hyperparameters are meticulously tuned to achieve optimal performance. We treat fake news detection as a binary classification problem and our evaluation criteria include F1, Precision, Recall, F1 Macro, and F1 Micro ~\cite{2022-get}. For more implementation details, see the
source code in this repository\footnote{\href{https://anonymous.4open.science/r/STEEL-6FD1/}{https://anonymous.4open.science/r/STEEL-6FD1/}}. Besides, cost details can be seen at \ref{cost details}.

\subsection{Main Results}


Our model, STEEL, was benchmarked against 11 baseline approaches, comprising 7 evidence-based and 4 LLM-based methods. We classified these into two groups: G1 for evidence-based methods and G2 for LLM-based methods. Performance metrics are reported in Tables \ref{tab:performance_liar}, \ref{tab:performance_chef}, and \ref{tab:performance_politifact}.
Key observations from these results include the following. 

\begin{enumerate}[label={\arabic*)}]
  \item STEEL consistently outperforms state-of-the-art methods in three real-world datasets, with more than a $5\%$ increase in both F1-macro and F1-micro scores. This also underscores the model's superior detection capabilities.
  \item In a detailed evaluation, we measured the performance of STEEL in three key metrics: F1, Precision, and Recall, classifying real news as positive and fake news as negative. STEEL demonstrated superior performance on these indicators.
  \item STEEL surpasses all baselines in the detection of fake news, evidenced by improved detection metrics. For instance, on the LIAR dataset, we observed increases in F1 False, Precision False, and Recall False by $17.3\%$, $11.5\%$, and $18.2\%$, respectively. Comparable significant gains were noted on other data sets.
\end{enumerate}

The collective evidence affirms that STEEL is highly effective in detecting fake news, with significant advantages in both reasoning and accuracy.

\subsection{Internet Search Comparison Study} \label{Internet Search Comparison Study}


To evaluate the relative effectiveness of our research mechanism compared to other methods in terms of improving the quality of evidence retrieval, we conducted a comparative experiment. 
The results are presented in Table \ref{tab:re-search}. 
"Re-search" represents our proposed scheme.
The alternative methods used for comparison involve single searches.
"Direct search" denotes the scenario where claims are directly used as queries for evidence retrieval. "Search with Keywords" involves the extraction of key terms from the claims before searching. "Search after Paraphrase" entails paraphrasing the claim before searching.




\begin{table}[!t]
    \small
    \centering 
    
    \begin{tabular}{lccc}
    \toprule
    \multirow{2}{*}{Method} & \multicolumn{3}{c}{F1-Ma}\\
    \cmidrule(lr){2-4}
    &LIAR & CHEF & PolitiFact \\
    \midrule
    Direct Search & 0.695 & 0.771 & 0.733 \\
    Search with Keywords & 0.699 & 0.775 & 0.735 \\
    Search after Paraphrase & 0.702 & 0.780 & 0.736 \\
    \midrule
    Re-search & \textbf{0.714} & \textbf{0.793} & \textbf{0.751} \\
    \bottomrule
    \end{tabular}
    
\caption{Performance comparison of various search strategies.}
\label{tab:re-search}
\end{table}

The results indicate that while certain conventional retrieval optimization methods employed by search engines, including keyword search and paraphrasing, offer improvements over the straightforward use of the claim as a query, their effectiveness remains notably inferior to that of the re-search module. 
This discrepancy arises from the fact that evidence obtained in a single search is insufficient to make a conclusive judgment. The results illustrate the important role of the re-search module in our framework.

\begin{table}[t]
     \centering
     \resizebox{0.48\textwidth}{!}{
     \begin{tabular}{llcccc}
     \toprule
     & \multicolumn{5}{c}{F1-Ma} \\
     \cline{2-6}
     & & $l=3$ & $l=4$ & $l=5$ & $l=all$ \\
     
     \multirow{4}{*}{LIAR} & $k=1$ & 0.631 & 0.636 & 0.643 & 0.671 \\
     & $k=3$ & 0.650 & 0.662 & 0.677 & \textbf{0.714} \\
     & $k=5$ & 0.673 & 0.675 & 0.684 & 0.713 \\
     & $k=7$ & 0.669 & 0.672 & 0.680 & 0.713 \\     
     \bottomrule
     \end{tabular}
     }
     \caption{The impact of the number of URLs ($k$) and paragraphs per document ($l$) on performance.}
     \label{tab:ret_con}
\end{table}

     
\subsection{Optimal Parameters in Evidence Selection}
To enhance the quality of evidence post-retrieval, we experimented with two key parameters, the number of document links ($k$) and length of the evidence ($l$). As shown in Table \ref{tab:ret_con}, the most significant improvement was achieved when $k=3$ and $l=all$. This aligns with our expectation that comprehending and reasoning about a statement benefit from comprehensive and detailed information compared to fragmented or limited snippets.

\subsection{Ablation Study and Sensitivity Analysis}
\begin{figure}[t]
    \centering
    \includegraphics[width=0.4\textwidth]{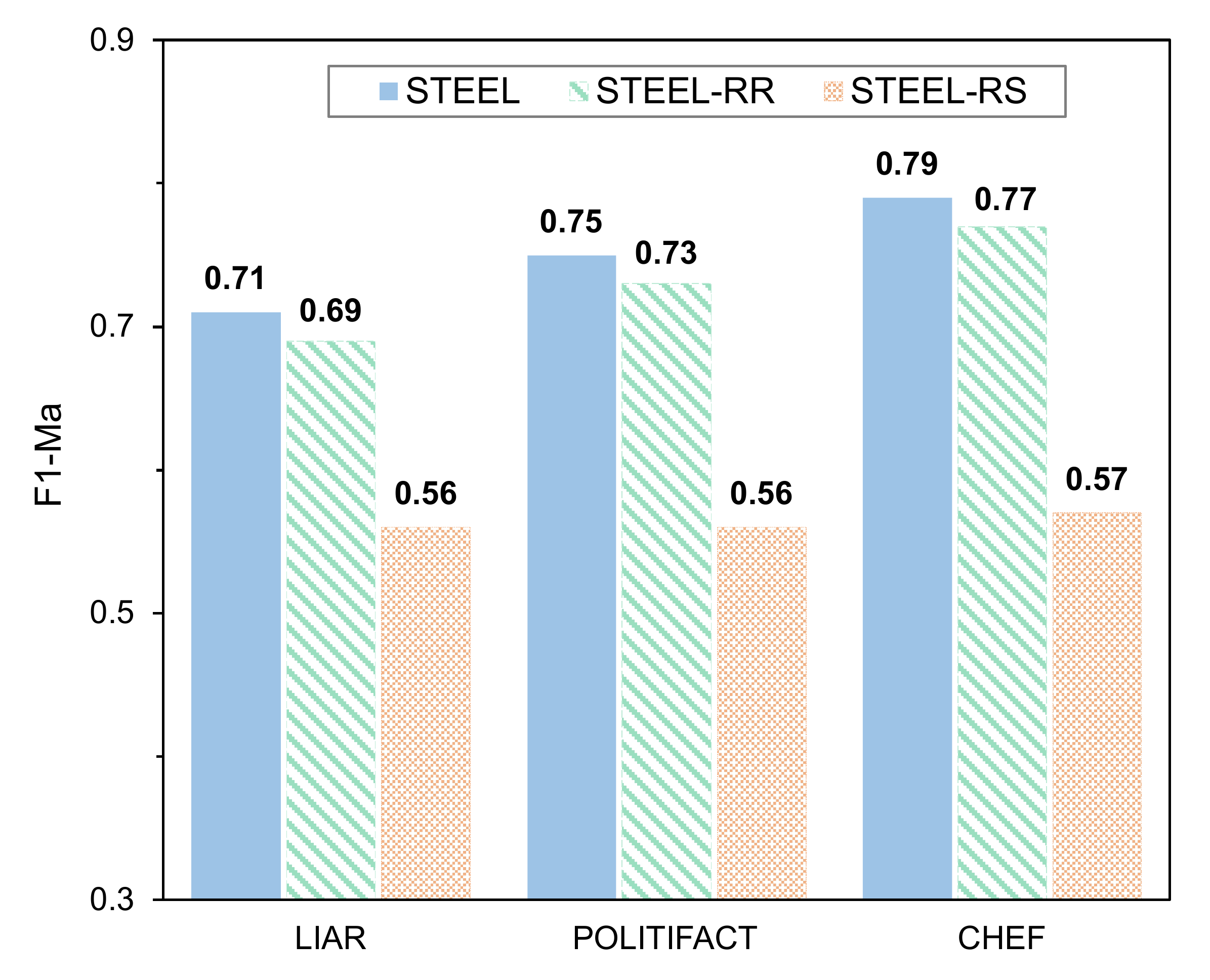}
    \caption{Ablation study results: STEEL denotes complete model performance, STEEL-RR represents removal of the re-search mechanism, and STEEL-RS represents GPT-3.5-Turbo without the search module.}
    \label{fig:abalation}
\end{figure}

In this section, we evaluated the impact of each module within our framework through ablation studies, as shown in Figure \ref{fig:abalation}. The label "-RS" marks the version of the model without the retrieval module, relying solely on LLMs, while "-RR" indicates the removal of the re-search mechanism. The results conclusively show that omitting any single component leads to diminished performance, validating that each module is integral to the framework's overall effectiveness. Notably, the retrieval module proves to be particularly critical for performance improvement. By effectively retrieving key evidence pertinent to the claim in question, the STEEL is better positioned to make accurate predictions, thereby highlighting the module's crucial role. Moreover, we carried out an ablation study on the semantic search module, with the results presented in Table \ref{tab:ssm_ablation}.
\begin{table}[t]
    \scriptsize
    \centering

    \resizebox{0.48\textwidth}{!}{
    \begin{tabular}{lccc}
    \toprule
       \multirow{2}{*}{Method}  & \multicolumn{3}{c}{F1-Ma}\\
    \cmidrule(lr){2-4}
    & LIAR & CHEF & PolitiFact \\
      \hline  
      Vanilla & 0.69 & 0.75 & 0.73 \\
      Quadratic Answer \cite{DBLP:journals/corr/abs-2308-07308} & 0.69 & 0.76 & 0.74  \\
      Response Correction \cite{DBLP:journals/corr/abs-2311-09000} & 0.69 & 0.75 & 0.74  \\
      Chain of Thought \cite{DBLP:conf/iclr/0002WSLCNCZ23} & 0.69 & 0.76 & 0.74  \\
      \midrule
      Round Control (STEEL) & \textbf{0.71} & \textbf{0.79} &  \textbf{0.75}\\
    \bottomrule
    \end{tabular}
    }
    \caption{Comparison of experimental results between round control and other methods}
    \label{tab:Prompt_effect}
\end{table}

Direct LLMs can not consistently achieve superior results, it is imperative to recognize the indispensable role of incorporating external evidence to substantially enhance predictive accuracy. \lgh{\begin{table}[!hp]
  \centering
  \small
\renewcommand{\arraystretch}{1.2}  
\begin{tabular}{lccc}

\toprule
\multirow{2}{*}{Method}  & \multicolumn{3}{c}{F1-Ma}\\
    \cmidrule(lr){2-4}
    & LIAR & CHEF & PolitiFact \\
      \hline  
STEEL w/o SSM & 0.702 & 0.780 & 0.739\\ \hline
STEEL & \textbf{0.714} & \textbf{0.793} & \textbf{0.751}\\ \hline
\end{tabular}
\caption{Experimental results of the ablation analysis of the semantic search module.}
  \label{tab:ssm_ablation}
\end{table}

}
The utilization of external evidence sources complements the limitations of LLMs and contributes to more robust and reliable predictions. Additionally, we conducted an ablation study to examine the impact of various step searches using different LLMs, with the findings detailed in Table \ref{tab:different_steps}.

\lgh{\begin{table}[!h]
\centering
\small
\renewcommand{\arraystretch}{1.2}
\resizebox{0.48\textwidth}{!}{
\begin{tabular}{lccc}
\hline
\multirow{2}{*}{Method}  & \multicolumn{3}{c}{F1-Ma}\\
    \cmidrule(lr){2-4}
    & LIAR & CHEF & POLITIFACT \\ 
\hline
Vicuna-7b & 0.528 & 0.519 & 0.522 \\
Vicuna-7b + BING 1-step search & 0.617 & 0.683 & 0.665 \\
Vicuna-7b + BING multi-stage search & 0.629 & 0.701 & 0.677 \\
\midrule
GPT-3.5-turbo & 0.563 & 0.574 & 0.567 \\
GPT-3.5-turbo + BING 1-step search & 0.691 & 0.770 & 0.738 \\
GPT-3.5-turbo + BING multi-stage search(STEEL) & \textbf{0.714} & \textbf{0.793} & \textbf{0.751} \\
\hline
\end{tabular}
}
\caption{Performence comparison of various combinations across three datasets.}
\label{tab:different_steps}
\end{table}}

\begin{figure}[!tbp]
	\centering
        \includegraphics[width=0.4\textwidth]{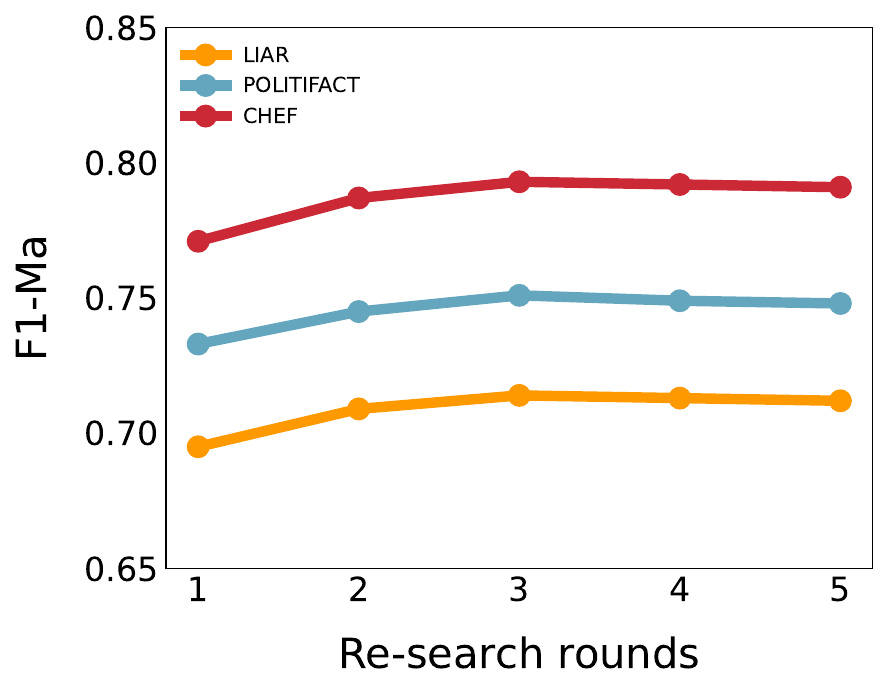}
    \caption{F1-Ma of various numbers of re-search rounds on three challenging claim verification datasets: LIAR (orange line), CHEF (red line), and PolitFact (blue line). }
    \label{fig:sen_ana}
\end{figure}

The primary insight gleaned from this experiment highlights that consistent performance improvement is achievable through the utilization of multiple retrieval steps. This implies that when the initial retrieval step fails to yield a relevant evidence passage, the retriever continues its efforts in subsequent iterative retrieval processes. As indicated by the results, the model's performance reaches its peak at 3 retrieval steps. Beyond this point, the addition of more steps does not yield substantial benefits and, in some cases, may even result in performance deterioration. Intriguingly, as shown in Figure \ref{fig:sen_ana}, the optimal number of retrieval steps remains consistent across datasets, regardless of variations in difficulty levels.




Since LLMs exhibit sensitivity to prompts, we have experimented with various kinds of prompts (examples of prompts can be seen in Appendix \ref{app:A.2}). The first category, denoted as "vanilla" prompts, encapsulates the unaltered combination of the original claim and the corresponding web pages retrieved from the Internet based on the claim. A second type, referred to as "quadratic answer" prompts\cite{DBLP:journals/corr/abs-2308-07308}, exclusively incorporates the explanatory text generated by the LLMs' initial output while omitting the answer component. These prompts are employed to engage the LLMs in self-consistency assessment. The third category, designated as "response correction" prompts\cite{DBLP:journals/corr/abs-2311-09000}, consists of both the explanatory content and the answer, facilitating a comprehensive evaluation of the coherence between the initial LLM output's response and its accompanying explanation. And the "Chain of Thought"\cite{DBLP:conf/iclr/0002WSLCNCZ23}, refers to denotes prompts that incorporate illustrative examples to guide LLMs in generating responses as per specific instructions. 



As shown in Table \ref{tab:Prompt_effect}, the results reveal that prompt adjustments often fail to improve, and can even detriment performance. Therefore, solely relying on prompt adjustments may not yield the desired outcomes, and our re-search provides a better approach regarding retrieving evidence.




\subsection{Explainability Study}

\begin{figure}
    \centering
    \includegraphics[width=0.48\textwidth]{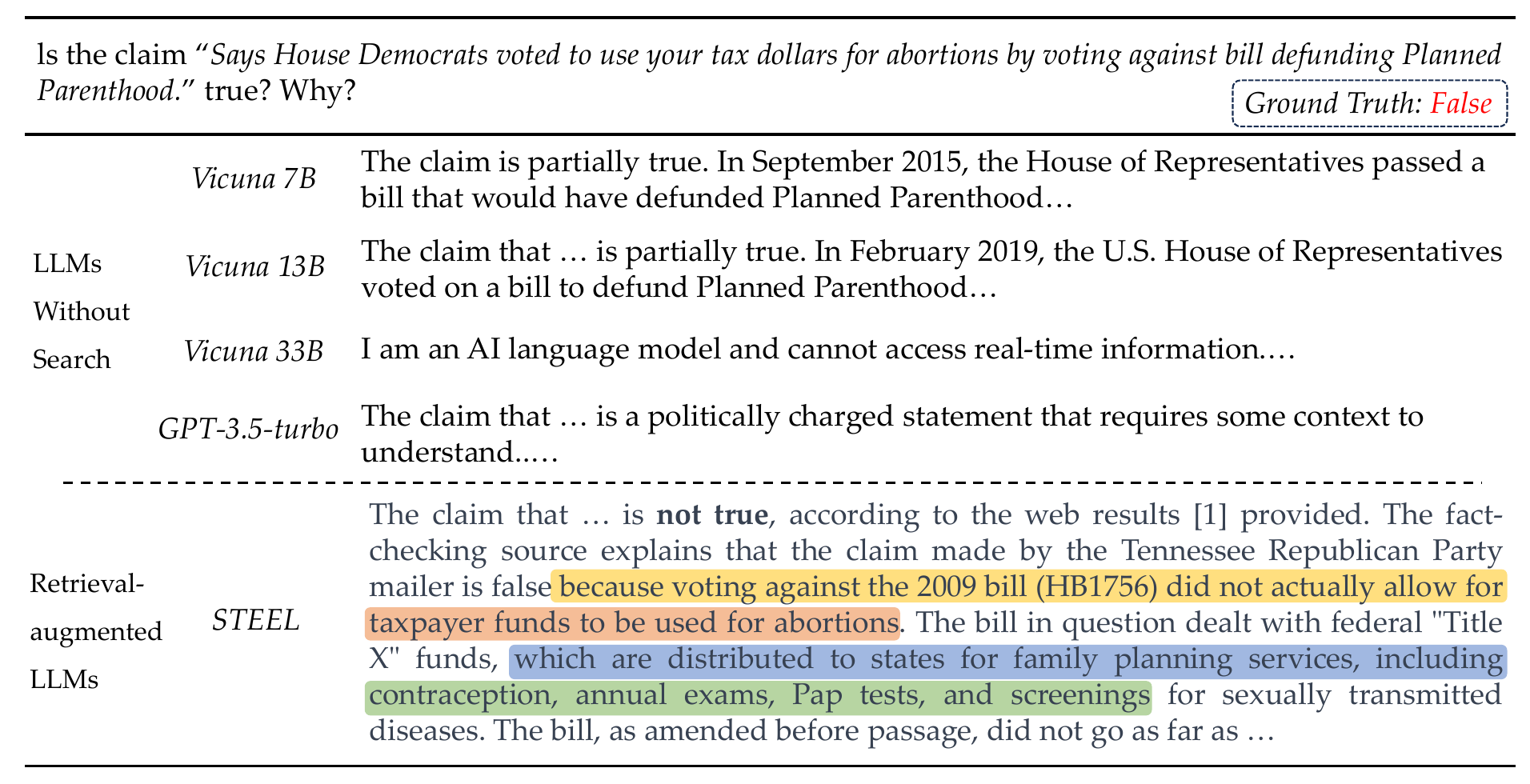}
    \caption{Explanation case study. Text with a colorful background indicates quoted evidence.}
    \label{fig:case}
\end{figure}

{\bf Case study}
In this section, we demonstrate the performance of our model in generating explanatory text. 
As shown in Figure \ref{fig:case}, we provide a specific example where a news claim asserted, "Says House Democrats voted to use your tax dollars for abortions by voting against bill defunding Planned Parenthood." Through the extraction of key evidence and coherent reasoning, our model effectively identified this news claim as false. More notably, our model is capable of reorganizing reasoning, utilizing complete evidence to craft human-friendly explanatory responses. Furthermore, it can attribute the generated text, distinguishing between factual information and generated content. This significantly enhances interpretability, benefiting both the model's understanding and the user's comprehension. 
\begin{table}[!pt]
  \centering
  \small 
    \begin{tabular}{ccccc}
    \toprule
    Method & F1-Ma & Precision & Agreement \\ \hline
    MUSER & 0.687 & 0.698 & 72.5\%   \\ \hline
    STEEL &	\textbf{0.773} & \textbf{0.741} & \textbf{78.2\%}  \\ 
    \bottomrule
    \end{tabular}
    \caption{Results of the user study. The agreement measure means the proportion of concurrence between the user's judgment and the model's judgment.}
     \label{tab:userstudy}
\end{table}

{\bf User study}
We assess whether real-world users can accurately discern the veracity of news claims using evidence obtained from STEEL. We selected $60$ claims from the CHEF and LIAR datasets, including $15$ authentic and $15$ false claims from each, and compared the quality of evidence provided by our STEEL model with that of MUSER. We hired $8$ college students to rate the evidence. To ensure methodological rigor, participants evaluated a randomized set of claims independently, without interaction. $10$ participants evaluated the evidence quality, reviewing either MUSER or STEEL-retrieved evidence for each claim and determining its truthfulness within a 3-minute timeframe. Participants also rated their confidence using a 5-point Likert scale. The results, depicted in Table \ref{tab:userstudy}, unequivocally demonstrate the superior performance of STEEL in evidence retrieval quality over MUSER.

\section{Conclusion}
In this paper, we present an out-of-the-box, end-to-end framework designed for fake news detection that centers around retrieval-augmented LLMs. Our work is a preliminary attempt to address systemic risks in the field of fake news detection, It has been proven that fully leveraging LLMs can aid individuals in identifying fake news by assisting in the gathering of ample evidence and facilitating judgment by end users. 
Considering the intricate challenges associated with identifying fake news, there is a significant need for the future to extend the framework's capabilities to encompass multimedia-based fake news, incorporating strategies to analyze and interpret information across text, images, videos, and audio. Addressing these areas will not only improve the accuracy and reliability of fake news detection but also broaden its applicability.

\paragraph{\textbf{Limitations}}  
Our study is constrained by two factors that warrant attention. A significant limitation of our methodology lies in the simplistic nature of the filtering algorithm utilized to identify fraudulent news sources. Currently, in the preprocessing of evidence, we employ a static blacklist to filter out recognized sources of disinformation. However, given the vast scale and rapid evolution of digital content, this approach may prove insufficient. We advocate for further investigation into this issue and the development of more advanced and diverse methods, including built-in mechanisms, for detecting and excluding counterfeit news outlets.

Additionally, the restricted context length of the input text poses another challenge, as it may not capture all relevant information adequately. This limitation underscores the need for additional research into the implications of context length restrictions within the domain of LLMs. Such exploration is essential for understanding their impact on efficacy and for identifying viable strategies for improvement.

Moreover, the technical quality of our method is hampered by the limited computational power available for fine-tuning current Large Language Models (LLMs). Nevertheless, we present a novel approach using existing LLMs with retrieval techniques for fake news detection, thereby laying the groundwork for future research endeavors.

\bibliography{custom}

\begin{thebibliography}{46}
\expandafter\ifx\csname natexlab\endcsname\relax\def\natexlab#1{#1}\fi

\bibitem[{Asai et~al.(2023)Asai, Min, Zhong, and Chen}]{DBLP:conf/acm/AsaiMZC23}
Akari Asai, Sewon Min, Zexuan Zhong, and Danqi Chen. 2023.
\newblock \href {https://doi.org/10.18653/v1/2023.acl-tutorials.6} {Retrieval-based language models and applications}.
\newblock In \emph{Proceedings of the 61st Annual Meeting of the Association for Computational Linguistics: Tutorial Abstracts, {ACL} 2023, Toronto, Canada, July 9-14, 2023}, pages 41--46. Association for Computational Linguistics.

\bibitem[{Borgeaud et~al.(2022)Borgeaud, Mensch, Hoffmann, Cai, Rutherford, Millican, van~den Driessche, Lespiau, Damoc, Clark, de~Las~Casas, Guy, Menick, Ring, Hennigan, Huang, Maggiore, Jones, Cassirer, Brock, Paganini, Irving, Vinyals, Osindero, Simonyan, Rae, Elsen, and Sifre}]{DBLP:conf/icml/BorgeaudMHCRM0L22}
Sebastian Borgeaud, Arthur Mensch, Jordan Hoffmann, Trevor Cai, Eliza Rutherford, Katie Millican, George van~den Driessche, Jean{-}Baptiste Lespiau, Bogdan Damoc, Aidan Clark, Diego de~Las~Casas, Aurelia Guy, Jacob Menick, Roman Ring, Tom Hennigan, Saffron Huang, Loren Maggiore, Chris Jones, Albin Cassirer, Andy Brock, Michela Paganini, Geoffrey Irving, Oriol Vinyals, Simon Osindero, Karen Simonyan, Jack~W. Rae, Erich Elsen, and Laurent Sifre. 2022.
\newblock \href {https://proceedings.mlr.press/v162/borgeaud22a.html} {Improving language models by retrieving from trillions of tokens}.
\newblock In \emph{International Conference on Machine Learning, {ICML} 2022, 17-23 July 2022, Baltimore, Maryland, {USA}}, volume 162 of \emph{Proceedings of Machine Learning Research}, pages 2206--2240. {PMLR}.

\bibitem[{Capuano et~al.(2023)Capuano, Fenza, Loia, and Nota}]{DBLP:journals/ijon/CapuanoFLN23}
Nicola Capuano, Giuseppe Fenza, Vincenzo Loia, and Francesco~David Nota. 2023.
\newblock \href {https://doi.org/10.1016/j.neucom.2023.02.005} {Content-based fake news detection with machine and deep learning: a systematic review}.
\newblock \emph{Neurocomputing}, 530:91--103.

\bibitem[{Chen and Shu(2023)}]{DBLP:journals/corr/abs-2311-05656}
Canyu Chen and Kai Shu. 2023.
\newblock \href {https://doi.org/10.48550/ARXIV.2311.05656} {Combating misinformation in the age of llms: Opportunities and challenges}.
\newblock \emph{CoRR}, abs/2311.05656.

\bibitem[{Chiang et~al.(2023)Chiang, Li, Lin, Sheng, Wu, Zhang, Zheng, Zhuang, Zhuang, Gonzalez et~al.}]{chiang2023vicuna}
Wei-Lin Chiang, Zhuohan Li, Zi~Lin, Ying Sheng, Zhanghao Wu, Hao Zhang, Lianmin Zheng, Siyuan Zhuang, Yonghao Zhuang, Joseph~E Gonzalez, et~al. 2023.
\newblock Vicuna: An open-source chatbot impressing gpt-4 with 90\%* chatgpt quality.
\newblock \emph{See https://vicuna. lmsys. org (accessed 14 April 2023)}.

\bibitem[{Collins et~al.(2021)Collins, Hoang, Nguyen, and Hwang}]{DBLP:journals/jiat/CollinsHNH21}
Botambu Collins, Dinh~Tuyen Hoang, Ngoc~Thanh Nguyen, and Dosam Hwang. 2021.
\newblock \href {https://doi.org/10.1080/24751839.2020.1847379} {Trends in combating fake news on social media - a survey}.
\newblock \emph{J. Inf. Telecommun.}, 5(2):247--266.

\bibitem[{Grover et~al.(2022)Grover, Angara, Akhtar, and Chakraborty}]{DBLP:conf/nips/GroverAA022}
Karish Grover, S.~M.~Phaneendra Angara, Md.~Shad Akhtar, and Tanmoy Chakraborty. 2022.
\newblock \href {http://papers.nips.cc/paper\_files/paper/2022/hash/3d57795f0e263aa69577f1bbceade46b-Abstract-Conference.html} {Public wisdom matters! discourse-aware hyperbolic fourier co-attention for social text classification}.
\newblock In \emph{NeurIPS}.

\bibitem[{Guu et~al.(2020{\natexlab{a}})Guu, Lee, Tung, Pasupat, and Chang}]{10.5555/3524938.3525306}
Kelvin Guu, Kenton Lee, Zora Tung, Panupong Pasupat, and Ming-Wei Chang. 2020{\natexlab{a}}.
\newblock Realm: Retrieval-augmented language model pre-training.
\newblock In \emph{Proceedings of the 37th International Conference on Machine Learning}, ICML'20. JMLR.org.

\bibitem[{Guu et~al.(2020{\natexlab{b}})Guu, Lee, Tung, Pasupat, and Chang}]{DBLP:conf/icml/GuuLTPC20}
Kelvin Guu, Kenton Lee, Zora Tung, Panupong Pasupat, and Ming{-}Wei Chang. 2020{\natexlab{b}}.
\newblock \href {http://proceedings.mlr.press/v119/guu20a.html} {Retrieval augmented language model pre-training}.
\newblock In \emph{Proceedings of the 37th International Conference on Machine Learning, {ICML} 2020, 13-18 July 2020, Virtual Event}, volume 119 of \emph{Proceedings of Machine Learning Research}, pages 3929--3938. {PMLR}.

\bibitem[{Helbling et~al.(2023)Helbling, Phute, Hull, and Chau}]{DBLP:journals/corr/abs-2308-07308}
Alec Helbling, Mansi Phute, Matthew Hull, and Duen~Horng Chau. 2023.
\newblock \href {https://doi.org/10.48550/ARXIV.2308.07308} {{LLM} self defense: By self examination, llms know they are being tricked}.
\newblock \emph{CoRR}, abs/2308.07308.

\bibitem[{Hu et~al.(2023)Hu, Hong, Guo, Wen, and Yu}]{DBLP:conf/sigir/HuHGWY23}
Xuming Hu, Zhaochen Hong, Zhijiang Guo, Lijie Wen, and Philip~S. Yu. 2023.
\newblock \href {https://doi.org/10.1145/3539618.3592049} {Read it twice: Towards faithfully interpretable fact verification by revisiting evidence}.
\newblock In \emph{Proceedings of the 46th International {ACM} {SIGIR} Conference on Research and Development in Information Retrieval, {SIGIR} 2023, Taipei, Taiwan, July 23-27, 2023}, pages 2319--2323. {ACM}.

\bibitem[{Izacard et~al.(2023)Izacard, Lewis, Lomeli, Hosseini, Petroni, Schick, Dwivedi{-}Yu, Joulin, Riedel, and Grave}]{DBLP:journals/jmlr/Gautier23}
Gautier Izacard, Patrick S.~H. Lewis, Maria Lomeli, Lucas Hosseini, Fabio Petroni, Timo Schick, Jane Dwivedi{-}Yu, Armand Joulin, Sebastian Riedel, and Edouard Grave. 2023.
\newblock \href {https://www.jmlr.org/papers/v24/23-0037.html} {Few-shot learning with retrieval augmented language models}.
\newblock \emph{J. Mach. Learn. Res.}, 24:251:1--251:43.

\bibitem[{Jiang et~al.(2023)Jiang, Xu, Gao, Sun, Liu, Dwivedi{-}Yu, Yang, Callan, and Neubig}]{DBLP:conf/emnlp/JiangXGSLDYCN23}
Zhengbao Jiang, Frank~F. Xu, Luyu Gao, Zhiqing Sun, Qian Liu, Jane Dwivedi{-}Yu, Yiming Yang, Jamie Callan, and Graham Neubig. 2023.
\newblock \href {https://aclanthology.org/2023.emnlp-main.495} {Active retrieval augmented generation}.
\newblock In \emph{Proceedings of the 2023 Conference on Empirical Methods in Natural Language Processing, {EMNLP} 2023, Singapore, December 6-10, 2023}, pages 7969--7992. Association for Computational Linguistics.

\bibitem[{Kandpal et~al.(2023)Kandpal, Deng, Roberts, Wallace, and Raffel}]{DBLP:conf/icml/KandpalDRWR23}
Nikhil Kandpal, Haikang Deng, Adam Roberts, Eric Wallace, and Colin Raffel. 2023.
\newblock \href {https://proceedings.mlr.press/v202/kandpal23a.html} {Large language models struggle to learn long-tail knowledge}.
\newblock In \emph{International Conference on Machine Learning, {ICML} 2023, 23-29 July 2023, Honolulu, Hawaii, {USA}}, volume 202 of \emph{Proceedings of Machine Learning Research}, pages 15696--15707. {PMLR}.

\bibitem[{Kotonya and Toni(2020)}]{DBLP:conf/coling/KotonyaT20}
Neema Kotonya and Francesca Toni. 2020.
\newblock \href {https://doi.org/10.18653/v1/2020.coling-main.474} {Explainable automated fact-checking: {A} survey}.
\newblock In \emph{Proceedings of the 28th International Conference on Computational Linguistics, {COLING} 2020, Barcelona, Spain (Online), December 8-13, 2020}, pages 5430--5443. International Committee on Computational Linguistics.

\bibitem[{Liao et~al.(2023)Liao, Peng, Huang, Zhang, Li, Shu, and Xie}]{DBLP:conf/kdd/LiaoPHZLSX23}
Hao Liao, Jiahao Peng, Zhanyi Huang, Wei Zhang, Guanghua Li, Kai Shu, and Xing Xie. 2023.
\newblock \href {https://doi.org/10.1145/3580305.3599873} {{MUSER:} {A} multi-step evidence retrieval enhancement framework for fake news detection}.
\newblock In \emph{Proceedings of the 29th {ACM} {SIGKDD} Conference on Knowledge Discovery and Data Mining, {KDD} 2023, Long Beach, CA, USA, August 6-10, 2023}, pages 4461--4472. {ACM}.

\bibitem[{Liu et~al.(2023)Liu, Lai, Yu, Xu, Zeng, Du, Zhang, Dong, and Tang}]{DBLP:conf/kdd/LiuLYXZDZDT23}
Xiao Liu, Hanyu Lai, Hao Yu, Yifan Xu, Aohan Zeng, Zhengxiao Du, Peng Zhang, Yuxiao Dong, and Jie Tang. 2023.
\newblock \href {https://doi.org/10.1145/3580305.3599931} {Webglm: Towards an efficient web-enhanced question answering system with human preferences}.
\newblock In \emph{Proceedings of the 29th {ACM} {SIGKDD} Conference on Knowledge Discovery and Data Mining, {KDD} 2023, Long Beach, CA, USA, August 6-10, 2023}, pages 4549--4560. {ACM}.

\bibitem[{Ma et~al.(2019)Ma, Gao, Joty, and Wong}]{DBLP:conf/acl/MaGJW19}
Jing Ma, Wei Gao, Shafiq~R. Joty, and Kam{-}Fai Wong. 2019.
\newblock \href {https://doi.org/10.18653/v1/p19-1244} {Sentence-level evidence embedding for claim verification with hierarchical attention networks}.
\newblock In \emph{Proceedings of the 57th Conference of the Association for Computational Linguistics, {ACL} 2019, Florence, Italy, July 28- August 2, 2019, Volume 1: Long Papers}, pages 2561--2571. Association for Computational Linguistics.

\bibitem[{Min et~al.(2022)Min, Rong, Bian, Xu, Zhao, Huang, and Ananiadou}]{DBLP:conf/www/MinRBXZHA22}
Erxue Min, Yu~Rong, Yatao Bian, Tingyang Xu, Peilin Zhao, Junzhou Huang, and Sophia Ananiadou. 2022.
\newblock \href {https://doi.org/10.1145/3485447.3512163} {Divide-and-conquer: Post-user interaction network for fake news detection on social media}.
\newblock In \emph{{WWW} '22: The {ACM} Web Conference 2022, Virtual Event, Lyon, France, April 25 - 29, 2022}, pages 1148--1158. {ACM}.

\bibitem[{Nakano et~al.(2021)Nakano, Hilton, Balaji, Wu, Ouyang, Kim, Hesse, Jain, Kosaraju, Saunders, Jiang, Cobbe, Eloundou, Krueger, Button, Knight, Chess, and Schulman}]{DBLP:journals/corr/abs-2112-09332}
Reiichiro Nakano, Jacob Hilton, Suchir Balaji, Jeff Wu, Long Ouyang, Christina Kim, Christopher Hesse, Shantanu Jain, Vineet Kosaraju, William Saunders, Xu~Jiang, Karl Cobbe, Tyna Eloundou, Gretchen Krueger, Kevin Button, Matthew Knight, Benjamin Chess, and John Schulman. 2021.
\newblock \href {http://arxiv.org/abs/2112.09332} {Webgpt: Browser-assisted question-answering with human feedback}.
\newblock \emph{CoRR}, abs/2112.09332.

\bibitem[{OpenAI(2022)}]{openai2022chatgpt}
OpenAI. 2022.
\newblock Chatgpt.
\newblock \url{https://chat.openai.com}.
\newblock Accessed: 2022-11-30.

\bibitem[{Pan et~al.(2023)Pan, Wu, Lu, Luu, Wang, Kan, and Nakov}]{DBLP:conf/acl/PanWLLWKN23}
Liangming Pan, Xiaobao Wu, Xinyuan Lu, Anh~Tuan Luu, William~Yang Wang, Min{-}Yen Kan, and Preslav Nakov. 2023.
\newblock \href {https://doi.org/10.18653/v1/2023.acl-long.386} {Fact-checking complex claims with program-guided reasoning}.
\newblock In \emph{Proceedings of the 61st Annual Meeting of the Association for Computational Linguistics (Volume 1: Long Papers), {ACL} 2023, Toronto, Canada, July 9-14, 2023}, pages 6981--7004. Association for Computational Linguistics.

\bibitem[{Papadogiannakis et~al.(2023)Papadogiannakis, Papadopoulos, Markatos, and Kourtellis}]{DBLP:conf/www/Papadogiannakis23}
Emmanouil Papadogiannakis, Panagiotis Papadopoulos, Evangelos~P. Markatos, and Nicolas Kourtellis. 2023.
\newblock \href {https://doi.org/10.1145/3543507.3583443} {Who funds misinformation? {A} systematic analysis of the ad-related profit routines of fake news sites}.
\newblock In \emph{Proceedings of the {ACM} Web Conference 2023, {WWW} 2023, Austin, TX, USA, 30 April 2023 - 4 May 2023}, pages 2765--2776. {ACM}.

\bibitem[{Popat et~al.(2018)Popat, Mukherjee, Yates, and Weikum}]{2018-declare}
Kashyap Popat, Subhabrata Mukherjee, Andrew Yates, and Gerhard Weikum. 2018.
\newblock \href {https://doi.org/10.18653/v1/D18-1003} {{D}e{C}lar{E}: Debunking fake news and false claims using evidence-aware deep learning}.
\newblock In \emph{Proceedings of the 2018 Conference on Empirical Methods in Natural Language Processing}, pages 22--32.

\bibitem[{Ram et~al.(2023)Ram, Levine, Dalmedigos, Muhlgay, Shashua, Leyton-Brown, and Shoham}]{ram-etal-2023-context}
Ori Ram, Yoav Levine, Itay Dalmedigos, Dor Muhlgay, Amnon Shashua, Kevin Leyton-Brown, and Yoav Shoham. 2023.
\newblock \href {https://doi.org/10.1162/tacl_a_00605} {In-context retrieval-augmented language models}.
\newblock \emph{Transactions of the Association for Computational Linguistics}, 11:1316--1331.

\bibitem[{Shi et~al.(2023)Shi, Min, Yasunaga, Seo, James, Lewis, Zettlemoyer, and Yih}]{DBLP:journals/corr/abs-2301-12652}
Weijia Shi, Sewon Min, Michihiro Yasunaga, Minjoon Seo, Rich James, Mike Lewis, Luke Zettlemoyer, and Wen{-}tau Yih. 2023.
\newblock \href {https://doi.org/10.48550/arXiv.2301.12652} {{REPLUG:} retrieval-augmented black-box language models}.
\newblock \emph{CoRR}, abs/2301.12652.

\bibitem[{Vo and Lee(2021)}]{2021-mac}
Nguyen Vo and Kyumin Lee. 2021.
\newblock Hierarchical multi-head attentive network for evidence-aware fake news detection.
\newblock In \emph{Proceedings of the 16th Conference of the European Chapter of the Association for Computational Linguistics: Main Volume.}, page 965–975.

\bibitem[{Wang and Shu(2023)}]{DBLP:conf/emnlp/WangS23a}
Haoran Wang and Kai Shu. 2023.
\newblock \href {https://aclanthology.org/2023.findings-emnlp.416} {Explainable claim verification via knowledge-grounded reasoning with large language models}.
\newblock In \emph{Findings of the Association for Computational Linguistics: {EMNLP} 2023, Singapore, December 6-10, 2023}, pages 6288--6304. Association for Computational Linguistics.

\bibitem[{Wang et~al.(2023{\natexlab{a}})Wang, Wei, Schuurmans, Le, Chi, Narang, Chowdhery, and Zhou}]{DBLP:conf/iclr/0002WSLCNCZ23}
Xuezhi Wang, Jason Wei, Dale Schuurmans, Quoc~V. Le, Ed~H. Chi, Sharan Narang, Aakanksha Chowdhery, and Denny Zhou. 2023{\natexlab{a}}.
\newblock \href {https://openreview.net/pdf?id=1PL1NIMMrw} {Self-consistency improves chain of thought reasoning in language models}.
\newblock In \emph{The Eleventh International Conference on Learning Representations, {ICLR} 2023, Kigali, Rwanda, May 1-5, 2023}. OpenReview.net.

\bibitem[{Wang et~al.(2023{\natexlab{b}})Wang, Li, Sun, and Liu}]{DBLP:conf/emnlp/WangLSL23}
Yile Wang, Peng Li, Maosong Sun, and Yang Liu. 2023{\natexlab{b}}.
\newblock \href {https://aclanthology.org/2023.findings-emnlp.691} {Self-knowledge guided retrieval augmentation for large language models}.
\newblock In \emph{Findings of the Association for Computational Linguistics: {EMNLP} 2023, Singapore, December 6-10, 2023}, pages 10303--10315. Association for Computational Linguistics.

\bibitem[{Wang et~al.(2023{\natexlab{c}})Wang, Reddy, Mujahid, Arora, Rubashevskii, Geng, Afzal, Pan, Borenstein, Pillai, Augenstein, Gurevych, and Nakov}]{DBLP:journals/corr/abs-2311-09000}
Yuxia Wang, Revanth~Gangi Reddy, Zain~Muhammad Mujahid, Arnav Arora, Aleksandr Rubashevskii, Jiahui Geng, Osama~Mohammed Afzal, Liangming Pan, Nadav Borenstein, Aditya Pillai, Isabelle Augenstein, Iryna Gurevych, and Preslav Nakov. 2023{\natexlab{c}}.
\newblock \href {https://doi.org/10.48550/ARXIV.2311.09000} {Factcheck-gpt: End-to-end fine-grained document-level fact-checking and correction of {LLM} output}.
\newblock \emph{CoRR}, abs/2311.09000.

\bibitem[{Wang et~al.(2023{\natexlab{d}})Wang, Mao, Wu, Ge, Wei, and Ji}]{DBLP:journals/corr/abs-2307-05300}
Zhenhailong Wang, Shaoguang Mao, Wenshan Wu, Tao Ge, Furu Wei, and Heng Ji. 2023{\natexlab{d}}.
\newblock \href {https://doi.org/10.48550/ARXIV.2307.05300} {Unleashing cognitive synergy in large language models: {A} task-solving agent through multi-persona self-collaboration}.
\newblock \emph{CoRR}, abs/2307.05300.

\bibitem[{Wei et~al.(2022{\natexlab{a}})Wei, Tay, Bommasani, Raffel, Zoph, Borgeaud, Yogatama, Bosma, Zhou, Metzler, Chi, Hashimoto, Vinyals, Liang, Dean, and Fedus}]{DBLP:journals/tmlr/WeiTBRZBYBZMCHVLDF22}
Jason Wei, Yi~Tay, Rishi Bommasani, Colin Raffel, Barret Zoph, Sebastian Borgeaud, Dani Yogatama, Maarten Bosma, Denny Zhou, Donald Metzler, Ed~H. Chi, Tatsunori Hashimoto, Oriol Vinyals, Percy Liang, Jeff Dean, and William Fedus. 2022{\natexlab{a}}.
\newblock \href {https://openreview.net/forum?id=yzkSU5zdwD} {Emergent abilities of large language models}.
\newblock \emph{Trans. Mach. Learn. Res.}, 2022.

\bibitem[{Wei et~al.(2022{\natexlab{b}})Wei, Wang, Schuurmans, Bosma, Ichter, Xia, Chi, Le, and Zhou}]{DBLP:conf/nips/Wei0SBIXCLZ22}
Jason Wei, Xuezhi Wang, Dale Schuurmans, Maarten Bosma, Brian Ichter, Fei Xia, Ed~H. Chi, Quoc~V. Le, and Denny Zhou. 2022{\natexlab{b}}.
\newblock \href {http://papers.nips.cc/paper\_files/paper/2022/hash/9d5609613524ecf4f15af0f7b31abca4-Abstract-Conference.html} {Chain-of-thought prompting elicits reasoning in large language models}.
\newblock In \emph{NeurIPS}.

\bibitem[{West and Bergstrom(2020)}]{West2020}
Jevin~D. West and Carl~T. Bergstrom. 2020.
\newblock \href {https://doi.org/10.1073/pnas.1912444117} {Misinformation in and about science}.
\newblock \emph{Proceedings of the National Academy of Sciences}.

\bibitem[{Wu et~al.(2023)Wu, Li, Deng, Xiong, and Hooi}]{DBLP:conf/cikm/WuLDXH23}
Jiaying Wu, Shen Li, Ailin Deng, Miao Xiong, and Bryan Hooi. 2023.
\newblock \href {https://doi.org/10.1145/3583780.3615015} {Prompt-and-align: Prompt-based social alignment for few-shot fake news detection}.
\newblock In \emph{Proceedings of the 32nd {ACM} International Conference on Information and Knowledge Management, {CIKM} 2023, Birmingham, United Kingdom, October 21-25, 2023}, pages 2726--2736. {ACM}.

\bibitem[{Wu et~al.(2020)Wu, Rao, Yang, Wang, and Nazir}]{2020-ehian}
Lianwei Wu, Yuan Rao, Xiong Yang, Wanzhen Wang, and Ambreen Nazir. 2020.
\newblock \href {https://doi.org/10.24963/ijcai.2020/193} {Evidence-aware hierarchical interactive attention networks for explainable claim verification}.
\newblock In \emph{Proceedings of the Twenty-Ninth International Joint Conference on Artificial Intelligence, {IJCAI-20}}, pages 1388--1394.

\bibitem[{Xiong et~al.(2023)Xiong, Hu, Lu, Li, Fu, He, and Hooi}]{DBLP:journals/corr/abs-2306-13063}
Miao Xiong, Zhiyuan Hu, Xinyang Lu, Yifei Li, Jie Fu, Junxian He, and Bryan Hooi. 2023.
\newblock \href {https://doi.org/10.48550/arXiv.2306.13063} {Can llms express their uncertainty? an empirical evaluation of confidence elicitation in llms}.
\newblock \emph{CoRR}, abs/2306.13063.

\bibitem[{Xu et~al.(2022)Xu, Wu, Liu, Wu, and Wang}]{2022-get}
Weizhi Xu, Junfei Wu, Qiang Liu, Shu Wu, and Liang Wang. 2022.
\newblock \href {https://doi.org/10.1145/3485447.3512122} {Evidence-aware fake news detection with graph neural networks}.
\newblock In \emph{Proceedings of the ACM Web Conference 2022}, page 2501–2510.

\bibitem[{Yao et~al.(2023{\natexlab{a}})Yao, Yu, Zhao, Shafran, Griffiths, Cao, and Narasimhan}]{DBLP:journals/corr/abs-2305-10601}
Shunyu Yao, Dian Yu, Jeffrey Zhao, Izhak Shafran, Thomas~L. Griffiths, Yuan Cao, and Karthik Narasimhan. 2023{\natexlab{a}}.
\newblock \href {https://doi.org/10.48550/arXiv.2305.10601} {Tree of thoughts: Deliberate problem solving with large language models}.
\newblock \emph{CoRR}, abs/2305.10601.

\bibitem[{Yao et~al.(2023{\natexlab{b}})Yao, Zhao, Yu, Du, Shafran, Narasimhan, and Cao}]{DBLP:conf/iclr/YaoZYDSN023}
Shunyu Yao, Jeffrey Zhao, Dian Yu, Nan Du, Izhak Shafran, Karthik~R. Narasimhan, and Yuan Cao. 2023{\natexlab{b}}.
\newblock \href {https://openreview.net/pdf?id=WE\_vluYUL-X} {React: Synergizing reasoning and acting in language models}.
\newblock In \emph{The Eleventh International Conference on Learning Representations, {ICLR} 2023, Kigali, Rwanda, May 1-5, 2023}. OpenReview.net.

\bibitem[{Ye and Durrett(2022)}]{DBLP:conf/nips/YeD22}
Xi~Ye and Greg Durrett. 2022.
\newblock \href {http://papers.nips.cc/paper\_files/paper/2022/hash/c402501846f9fe03e2cac015b3f0e6b1-Abstract-Conference.html} {The unreliability of explanations in few-shot prompting for textual reasoning}.
\newblock In \emph{NeurIPS}.

\bibitem[{Yu et~al.(2023)Yu, Iter, Wang, Xu, Ju, Sanyal, Zhu, Zeng, and Jiang}]{DBLP:conf/iclr/0002IWXJ000023}
Wenhao Yu, Dan Iter, Shuohang Wang, Yichong Xu, Mingxuan Ju, Soumya Sanyal, Chenguang Zhu, Michael Zeng, and Meng Jiang. 2023.
\newblock \href {https://openreview.net/pdf?id=fB0hRu9GZUS} {Generate rather than retrieve: Large language models are strong context generators}.
\newblock In \emph{The Eleventh International Conference on Learning Representations, {ICLR} 2023, Kigali, Rwanda, May 1-5, 2023}. OpenReview.net.

\bibitem[{Zhang and Ghorbani(2020)}]{DBLP:journals/ipm/ZhangG20}
Xichen Zhang and Ali~A. Ghorbani. 2020.
\newblock \href {https://doi.org/10.1016/j.ipm.2019.03.004} {An overview of online fake news: Characterization, detection, and discussion}.
\newblock \emph{Inf. Process. Manag.}, 57(2):102025.

\bibitem[{Zhou and Zafarani(2020)}]{zhou2020survey}
Xinyi Zhou and Reza Zafarani. 2020.
\newblock A survey of fake news: Fundamental theories, detection methods, and opportunities.
\newblock \emph{ACM Computing Surveys (CSUR)}, 53(5):1--40.

\bibitem[{Zhou and Zafarani(2021)}]{DBLP:journals/csur/ZhouZ20}
Xinyi Zhou and Reza Zafarani. 2021.
\newblock \href {https://doi.org/10.1145/3395046} {A survey of fake news: Fundamental theories, detection methods, and opportunities}.
\newblock \emph{{ACM} Comput. Surv.}, 53(5):109:1--109:40.

\end{thebibliography}
\bibliographystyle{acl_natbib}
\clearpage

\appendix
\section{Appendix}

\subsection{Cost details} \label{cost details}

\begin{enumerate}[label={\arabic*)}]
    \item Regarding implementation details, we put detailed prompts in the appendix, and both BING search and ChatGPT parameters are available in the open-source code files. 
    \item Based on rough estimation, the cost is approximately $0.04$ dollars per test, among them, gpt-3.5-turbo costs $0.0015$ dollars per 1k tokens, and Bing search costs $15$ dollars per $1k$ searches. 
    \item We aim to provide an open-source, user-friendly fake news
detection framework with improved performance and interpretability. This framework can help people automatically verify a piece of news when they can’t decide.
\end{enumerate}

\subsection{Baselines' Description}\label{app: baselines}

\setcounter{table}{0}
\setcounter{figure}{0}
\renewcommand{\thetable}{A\arabic{table}}
\renewcommand{\thefigure}{A\arabic{figure}}



\textbf{Evidence-based methods}
\begin{itemize}
    \item \textbf{DeClarE} (EMNLP’18)~\cite{2018-declare}: They employ BiLSTM (Bidirectional Long Short-Term Memory) to embed the semantics of evidence and calculate evidence scores via an attention interaction mechanism.
    
    \item \textbf{HAN} (ACL'19) \cite{DBLP:conf/acl/MaGJW19} HAN utilizes GRU (Gated Recurrent Unit ) embedding and incorporates two modules: one for topic consistency and another for semantic entailment. These modules are based on a sentence-level attention mechanism, facilitating the simulation of claim-evidence interaction.

    \item \textbf{EHIAN} (IJCAI’20)~\cite{2020-ehian}: EHIAN achieves interpretable claim verification by employing an evidence-aware hierarchical interactive attention network, enabling the exploration of  more plausible evidence semantics.
    

    \item \textbf{MAC} (ACL’21)~\cite{2021-mac}: MAC combines multi-head word-level attention and multi-head document-level attention, facilitating the interpretation for fake news detection at both word-level and evidence-level.

    \item \textbf{GET} (WWW’22)~\cite{2022-get}: GET models claims and pieces of evidence as graph-structured data, allowing for the exploration of complex semantic structures. Furthermore, it mitigates information redundancy through the incorporation of a semantic structure refinement layer.

    \item \textbf{MUSER} (KDD'23)~\cite{DBLP:conf/kdd/LiaoPHZLSX23}: MUSER adopts a multi-step evidence retrieval strategy, leveraging the interdependence among multiple pieces of evidence to enhance its performance.

    \item \textbf{ReRead} (SIGIR'23)~\cite{DBLP:conf/sigir/HuHGWY23} : ReRead retrieves appropriate evidence from real-world documents by applying standards of plausibility, sufficiency, and sufficiency. 

\end{itemize}

\textbf{LLM-based methods with or without retrieval}
\begin{itemize}
    \item \textbf{GPT-3.5-turbo}~\cite{openai2022chatgpt}: ChatGPT is a sibling model to InstructGPT, and it shares similarities with InstructGPT in terms of its capacity to understand and respond to prompts with detailed explanations. The baseline model employed here is gpt-3.5-turbo.    
    \item \textbf{Vicuna-7B}~\cite{chiang2023vicuna}: Vicuna is a fully open-source base model for Large Language Models (LLMs) that has gained widespread usage. It is trained through the process of fine-tuning LLaMA on conversational data shared by users, which was collected from ShareGPT.
    \item \textbf{WEBGLM} (KDD'23)~\cite{DBLP:conf/kdd/LiuLYXZDZDT23} : WEBGLM is a web-enhanced question-answering system based on the General Language Model (GLM). It retrieves relevant content from the Internet and then feeds it into LLMs for analysis. The baseline model utilized here is the 2B version with Bing search integration.
    \item \textbf{ProgramFC} (ACL'23)~\cite{DBLP:conf/acl/PanWLLWKN23} : ProgramFC is a fact-checking model that decomposes complex claims into simpler sub-claims that can be solved with a shared library of specialized functions. It uses strategic retrieval powered by Codex for fact checking. The baseline setting we employ is their open-book configuration.
\end{itemize}

\begin{figure}[t]
    \centering
    \includegraphics[width=0.5\textwidth]{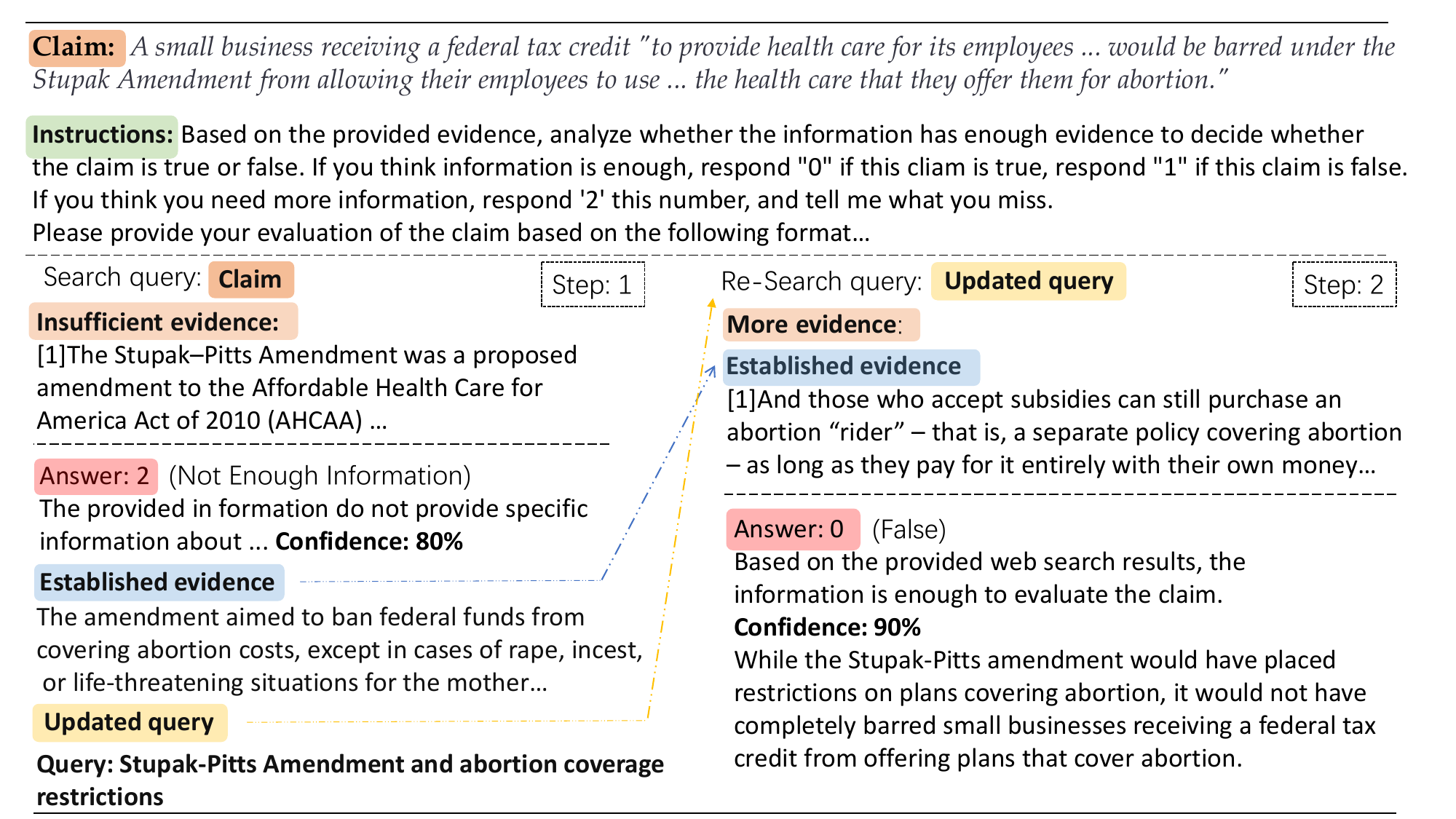}
    \caption{
    LLMs can make decisions based on given evidence, including deciding if they need to re-search.
    } 
    \label{fig:hypo}
\end{figure}

     

\section{Prompts' Description}
\subsection{Type Description}

\label{app:A.2}

\begin{itemize}
    \item \textbf{Compress Prompt:} Initial search may lack sufficient data to verify the claim, but the retrieved evidence can still guide subsequent searches. These prompts combine the original claim with newly retrieved information, feeding them into LLMs for compressed data
    \item \textbf{Keywords Prompt:} This prompt is used to extract several keywords from claim and then retrieve. This is actually STEEL with a 1 step number, but the prompt at the beginning has been changed to keywords.
    \item \textbf{Paraphrase Prompt:} This prompt is used to paraphrase original claim into another sentence and then retrieve. This is actually STEEL with 1 step, but the prompt at the beginning has been changed to sentence after paraphrased.
    \item \textbf{Quadratic answer:} This prompt is used to check whether the explanation text output by the LLMs model is consistent. The actual purpose here is to try to solve the problem of self-contradiction in the answers output by LLMs models. However, the effect is not ideal.
    \item \textbf{Response correction:} This prompt aims to verify the consistency of LLMs' complete output and address the issue of self-contradictory answers produced by these models. 
    \item \textbf{Step check:} This prompt is used as part of our STEEL model. 
    \item \textbf{Text summary:} This prompt is used to convert long news into verifiable short claims.
    \item \textbf{Vanilla:} This prompt is used in one-shot retrieval-enhanced LLMs.
\end{itemize} 

\definecolor{backcolour}{rgb}{0.96, 0.96, 0.96}
\definecolor{codegreen}{rgb}{0,0.6,0}
\lstdefinestyle{myStyle}{
    backgroundcolor=\color{backcolour},   
    commentstyle=\color{codegreen},
    basicstyle=\ttfamily\small,
    breakatwhitespace=true,         
    breaklines=true,                 
    keepspaces=true,                 
    numbers=none,       
    numbersep=5pt,                  
    showspaces=false,                
    showstringspaces=false,
    showtabs=false,                  
    tabsize=2,
    frame=single,
}
\lstset{style=myStyle}

\onecolumn
\subsection{Prompts Examples}
\label{sec: prompts}

\begin{lstlisting}[caption=Compress Prompt, label={lst:compress}]
Instruction: The information obtained from this search is not enough to determine whether the claim is true or false, so I need you to help me extract key evidence from the information obtained this time and compress it into one or two sentences for subsequent use.
Statement: {search_text} 
Web search results: {context_str}
Compressed information: [Compressed information]
\end{lstlisting}

\begin{lstlisting}[caption=Keywords Prompt, label={lst:keywords}]
Instruction: Given a claim, if I want to verify the truth or falseness of the claim, help me extract the keywords of the claim to be more suitable for web search engines to search for evidence. The keywords should be short and no more than 4.
Statement: {search_text}
Keywords: ["Keyword 1", "Keyword 2"...]
Remember to follow the format for output.
\end{lstlisting}

\begin{lstlisting}[caption=Paraphrase Prompt, label={lst:paraphrase}]
Instructions: Given a claim, if I want to verify the truth or falsehood of the claim, help me paraphrase the statement so that it is more suitable for web search engines to search for evidence. The paraphrase should be one sentence.
Claim: {search_text}
Paraphrased: [paraphrased text]
Remember to follow the format for output.
\end{lstlisting}

\begin{lstlisting}[caption=Quadratic answer(qa) Prompt, label={lst:quadratic_answer}]
Instruction: I will provide you with a paragraph that contains a judgment on a claim.
Your task is to analyze the intention of this paragraph. If the paragraph considers the claim to be true, reply with 'Answer: 0'. If the paragraph considers the statement to be false, reply with 'Answer: 1'. If the paragraph considers the statement to be neither true nor false, reply with 'Answer: 2'.
Paragraph: {first_response_text}
Answer: [0/1/2]
Please remember to follow my instructions to reply.
\end{lstlisting}


\begin{lstlisting}[caption=Response correction(rc) Prompt, label={lst:response_coorection}]
Instructions: I will give you the following information: Explanation: [Explain why you make this judgment.]
Answer: [0/1/2]
If Answer is 0 or 1: Just provide the number.
If Answer is 2: 
Missing info: [Description of missing information]
Query: ["Query 1", "Query 2", ...] 
Your task is to help me determine whether the number after "Answer: " is consistent with the explanation, where "Answer: 0" means this cliam is true , "Answer: 1" means this claim is false. "Answer: 2" means you need more information. If they are inconsistent, the explanation shall prevail and the Answer shall be corrected, and the answer with only Answer: changed shall be returned.
The paragraph that needs to be judged and changed: {first_response_text}
\end{lstlisting}

\begin{lstlisting}[caption=One-shot Prompt, label={lst:one_shot}]
Instructions: Based on the provided web search results, analyze whether the information has enough evidence to decide whether the statement is true or false. If you think information is enough, respond "0" if this cliam is true, respond "1" if this claim is false. If you think you need more information, respond '2' this number, and tell me what you miss and what should you search. Please provide your evaluation of the claim based on the following format: Answer: [0/1/2] If Answer is 0 or 1: Just provide the number If Answer is 2:
Missing info: [Description of missing information]
Query: ["Query 1", "Query 2", ...]

Here is an example:Q: Instructions: Based on the provided web search results, analyze whether the information has enough evidence to decide whether the statement is true or false.
Claim: This is rob-portman's statement. Under Lt. Gov. Lee Fisher, Ohio is 44th in the country in terms of getting money actually into worker retraining.
Web search result:
Source [1] cleveland.com
"Under Lt. Gov. Lee Fisher, Ohio is 44th in the country in terms of getting money actually into worker retraining ..."
Source [2] politifact.com
"I agree with what Joe Hallett just said. Sometimes trade has a disruptive effect and we need to be sure that we are minimizing that. But when it happens, we need to be sure that the government steps in and provides ..."
Please provide your evaluation of the claim based on the following format:
Answer: 2
Missing info: The sources do not provide enough context about what exactly is being ranked 44th. They also do not conclusively state whether the 44th ranking was current at the time the claim was made.
Query: ["Ohio worker retraining funding rankings over time", "Ohio worker retraining program performance under Lee Fisher"]

Now is your turn:
Claim: This is dan-pfeiffer's statement. Says President Barack Obamas approval rating gained 3 points in the last couple months.Web search result:Source [1] politifact.com
When asked about Obama's approval rating, Pfeiffer said, "...Remember to fill in all required fields based on the Answer value.If you think information is enough, respond "0" if this claim is true, respond "1" if this claim is false.If you think you need more information, respond '2' this number, and tell me what you miss and what should you search in the format I specified.
\end{lstlisting}

\begin{lstlisting}[caption=Step check(sc) Prompt, label={lst:step_check}]
Instructions: Based on the provided web search results, analyze whether the information has enough evidence to decide whether the claim is true or false.
If you think information is enough, respond "0" 
if this claim is true, respond "1" if this claim is false.
If you think the claim is partially true, respond "0".
If you think you need more information, respond '2' this number, and tell me what you miss and what should you search; and I need you to help me extract key evidence from the information obtained this time and compress it into one or two sentences for subsequent use.
Please provide your evaluation of the claim based on the following format: 
Explanation: [Explain why you make this judgement.]
Answer: [0/1/2]
If Answer is 0 or 1: Just provide the number.
If Answer is 2: 
Missing info: [Description of missing information]
Query: ["Query 1", "Query 2", ...]
Compressed information: [Compressed information]
Claim: {search_text}
Web search result: {context_str}
Remember to fill in all required fields based on the Answer value. If you think information is enough, respond "0" if this claim is true, respond "1" if this claim is false. If you think you need more information, respond '2' this number, and tell me what you miss and what should you search in the format I specified. And explain how confident you are (0~100%) Confirm that you are confident in your answer, and reply "Confidence: [0~100%].
\end{lstlisting}

\begin{lstlisting}[caption=Text summary(ts) Prompt, label={lst:text_summary}]
Instructions: Please help me convert this news article into a concise claim that can be used to assess its authenticity.
Claim:[Key claim of news]
Remember, the claim should be brief and to the point. 
text: {news}
\end{lstlisting}

\begin{lstlisting}[caption=Vanilla(va) Prompt, label={lst:vanilla}] Instructions: Based on the provided web search results, analyze whether the claim is true, false.
Claim: {search_text} Web search result: {context_str} Answer: [true/false] Remember to fill in all required fields based on your judgement.
You must and can only choose one answer from true or false.
\end{lstlisting}




\end{document}